\documentclass[review]{elsarticle}

\usepackage{hyperref}
\usepackage{microtype}
\usepackage{graphicx}
\usepackage{booktabs} 
\usepackage{algorithm}
\usepackage{algorithmic}
\usepackage[table]{xcolor}
\usepackage{pgfplotstable}
\usepackage{nicefrac}       
\usepackage{microtype}      
\usepackage{bm}
\usepackage{bbm}
\usepackage{array,multirow}
\usepackage{float}
\usepackage{subfig}
\usepackage{comment}
\usepackage{amsmath}

\usepackage{amsfonts}

\usepackage{tikz}

\usepackage{pgfplots}
\usetikzlibrary{arrows}

\journal{Arxiv}

\begin{document}

\begin{frontmatter}

\title{Squeezing Backbone Feature Distributions to the Max for Efficient Few-Shot Learning}


\author[orange,imt]{Yuqing Hu}

\author[orange]{St\'ephane Pateux}
\author[imt]{Vincent Gripon}

\address[orange]{Orange Labs, Rennes, France}
\address[imt]{IMT Atlantique, Lab-STICC, UMR CNRS 6285, F-29238, France}

\begin{abstract}
Few-shot classification is a challenging problem due to the uncertainty caused by using few labelled samples. In the past few years, many methods have been proposed with the common aim of transferring knowledge acquired on a previously solved task, what is often achieved by using a pretrained feature extractor. Following this vein, in this paper we propose a novel transfer-based method which aims at processing the feature vectors so that they become closer to Gaussian-like distributions, resulting in increased accuracy. In the case of transductive few-shot learning where unlabelled test samples are available during training, we also introduce an optimal-transport inspired algorithm to boost even further the achieved performance. Using standardized vision benchmarks, we show the ability of the proposed methodology to achieve state-of-the-art accuracy with various datasets, backbone architectures and few-shot settings.
\end{abstract}


\end{frontmatter}


\section{Introduction}
\label{introduction}

Thanks to their outstanding performance, Deep Learning methods have been widely considered for vision tasks such as image classification and object detection. In order to reach top performance, these systems are typically trained using very large labelled datasets that are representative enough of the inputs to be processed afterwards.

However, in many applications, it is costly to acquire or to annotate data, resulting in the impossibility to create such large labelled datasets. Under this condition, it is challenging to optimize Deep Learning architectures considering the fact they typically are made of way more parameters than the dataset can efficiently tune. This is the reason why in the past few years, few-shot learning (i.e. the problem of learning with few labelled examples) has become a trending research subject in the field. In more details, there are two settings that authors often consider: a) ``inductive few-shot'', where only a few labelled samples are available during training and prediction is performed on each test input independently, and b) ``transductive few-shot'', where prediction is performed on a batch of (non-labelled) test inputs, allowing to take into account their joint distribution.

Many works in the domain are built based on a ``learning to learn'' guidance, where the pipeline is to train an optimizer~\cite{finn2017model, DBLP:conf/iclr/RaviL17, thrun2012learning} with different tasks of limited data so that the model is able to learn generic experience for novel tasks. Namely, the model learns a set of initialization parameters that are in an advantageous position for the model to adapt to a new (small) dataset. Recently, the trend evolved towards using well-thought-out transfer architectures (called backbones)~\cite{torrey2010transfer, DBLP:journals/tip/DasL20, DBLP:conf/iclr/ChenLKWH19, mangla2020charting, hu2021graph, hu2021leveraging} trained one time on the same training data, but seen as a unique large dataset.

A main problem of using features extracted using a backbone pretrained architecture is that their distribution is likely to be unorthodox, as the problem the backbone has been optimized for most of the time differs from that it is then used upon. As such, methods that rely on strong assumptions about the feature distributions tend to have limitations on leveraging their quality. In this paper, we propose an efficient feature preprocessing methodology that allows to boost the accuracy in few-shot transfer settings. In the case of transductive few-shot learning, we also propose an optimal transport based algorithm that allows reaching even better performance. Using standardized benchmarks in the field, we demonstrate the ability of the proposed method to obtain state-of-the-art accuracy, for various problems and backbone architectures.

\usetikzlibrary{shapes,arrows}
\tikzstyle{block} = [draw, fill=blue!20, rectangle, 
    minimum height=3em, minimum width=6em]
\tikzstyle{sum} = [draw, fill=blue!20, circle, node distance=1cm]
\tikzstyle{input} = [coordinate]
\tikzstyle{output} = [coordinate]
\tikzstyle{pinstyle} = [pin edge={to-,dashed, thin,black}]

\begin{figure*}
    \centering
    \begin{tikzpicture}[auto, node distance=2cm,>=latex', scale=0.5, every node/.style={scale=0.5}]
    \draw[fill=black,fill opacity=0.1,draw=black,draw opacity=0.1]
    (-8.2,0) rectangle (-4.2,6)
    (-0.2,0) rectangle (4.2,6)
    (6.2,0) rectangle (12.7,6);
    
    \draw[dashed, fill=orange,fill opacity=0.05, draw=black]
    (-10.2,-0.10) rectangle (-4.10,6.55);
    \node[text width=50mm, text centered, rotate=90] at (-9.25, 3.25) {\large Offline training of a generic feature extractor using a large available dataset};
    \draw[dashed, fill=green,fill opacity=0.05, draw=black]
    (-3.5,-0.10) rectangle (12.80,6.55);
    \node[text width=40mm, text centered, rotate=90] at (-1.90, 2.60) {\large Proposed PEME-BMS to learn to classify the considered few-shot dataset};
    
    \draw[]
    (-8.25,-0.05) rectangle (-4.15,6.5)
    (-0.25,-0.05) rectangle (4.25,6.5)
    (6.15,-0.05) rectangle (12.75,6.5);
    \node at (-6.2,6.25) {\textbf{Feature extraction}};
    \node at (2,6.25) {\textbf{Preprocessing}};
    \node at (9.4,6.25) {\textbf{Boosted Min-size Sinkhorn}};
    
    \node at (-6.2,5.5) {large dataset $\mathbf{D}_{base}$};
    \begin{scope}[xshift=-8.1cm, yshift=1.5cm]
    \setlength{\fboxsep}{1pt}
    \node at (1,3) {\fcolorbox{black}{white}{\includegraphics[width=1cm,height=1cm]{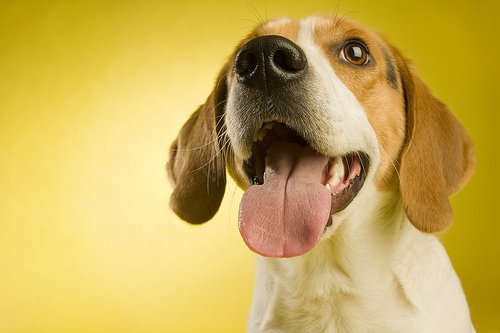}}};
    \node at (0.95,2.95) {\fcolorbox{black}{white}{\includegraphics[width=1cm,height=1cm]{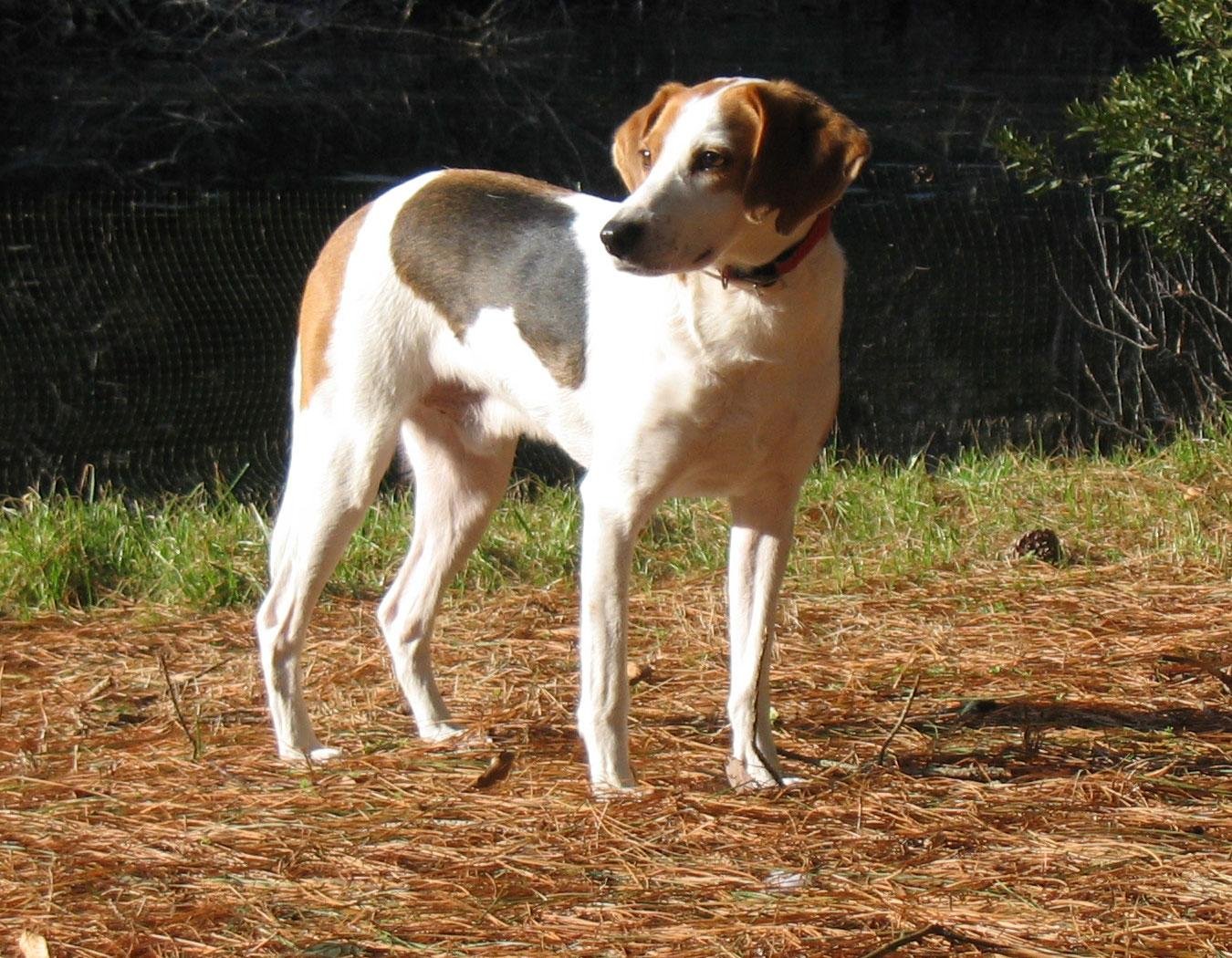}}};
    \node at (0.90,2.90) {\fcolorbox{black}{white}{\includegraphics[width=1cm,height=1cm]{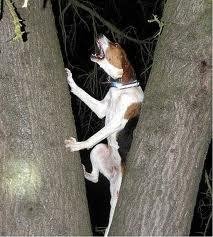}}};
    \node at (0.85,2.85) {\fcolorbox{black}{white}{\includegraphics[width=1cm,height=1cm]{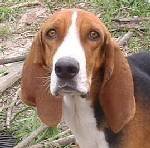}}};
    \node at (0.80,2.80) {\fcolorbox{black}{white}{\includegraphics[width=1cm,height=1cm]{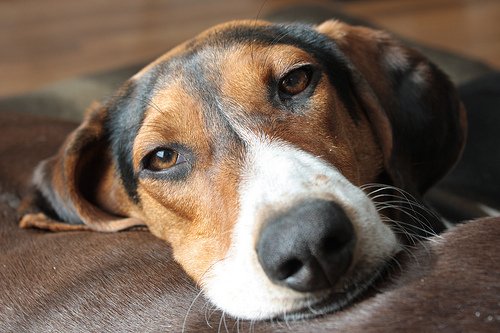}}};
    \node at (3,3) {\fcolorbox{black}{white}{\includegraphics[width=1cm,height=1cm]{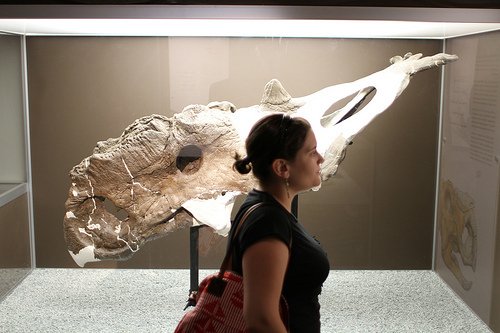}}};
    \node at (2.95,2.95) {\fcolorbox{black}{white}{\includegraphics[width=1cm,height=1cm]{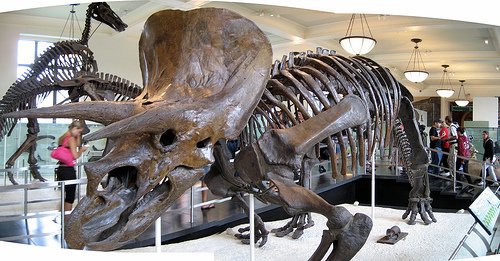}}};
    \node at (2.90,2.90) {\fcolorbox{black}{white}{\includegraphics[width=1cm,height=1cm]{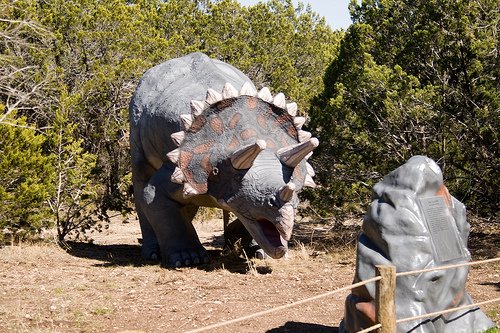}}};
    \node at (2.85,2.85) {\fcolorbox{black}{white}{\includegraphics[width=1cm,height=1cm]{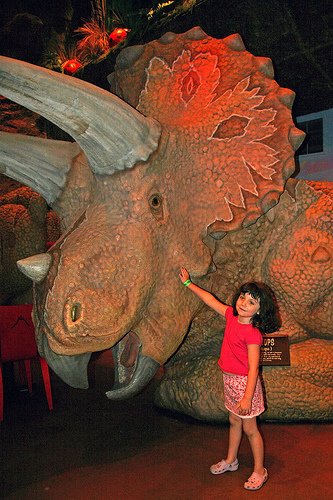}}};
    \node at (2.80,2.80) {\fcolorbox{black}{white}{\includegraphics[width=1cm,height=1cm]{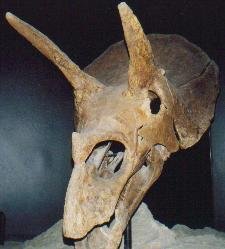}}};
    \node at (2,1.7) {\fcolorbox{black}{white}{\includegraphics[width=1cm,height=1cm]{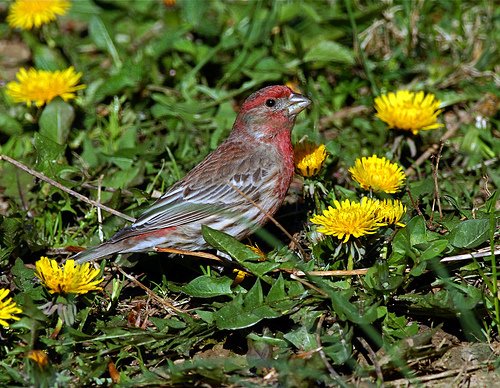}}};
    \node at (1.95,1.65) {\fcolorbox{black}{white}{\includegraphics[width=1cm,height=1cm]{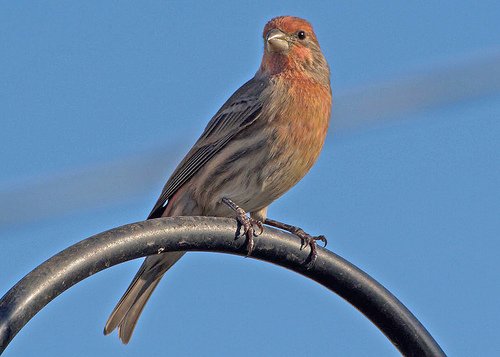}}};
    \node at (1.90,1.60) {\fcolorbox{black}{white}{\includegraphics[width=1cm,height=1cm]{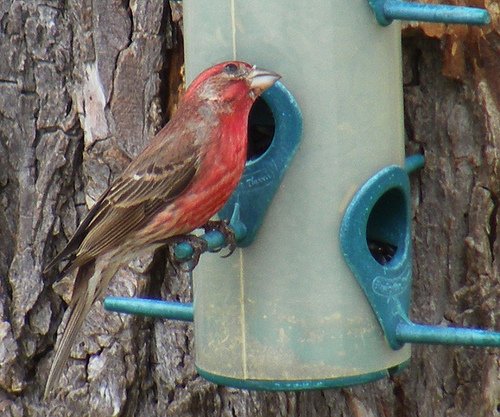}}};
    \node at (1.85,1.55) {\fcolorbox{black}{white}{\includegraphics[width=1cm,height=1cm]{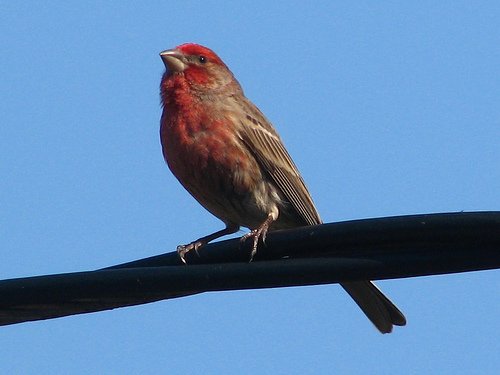}}};
    \node at (1.80,1.50) {\fcolorbox{black}{white}{\includegraphics[width=1cm,height=1cm]{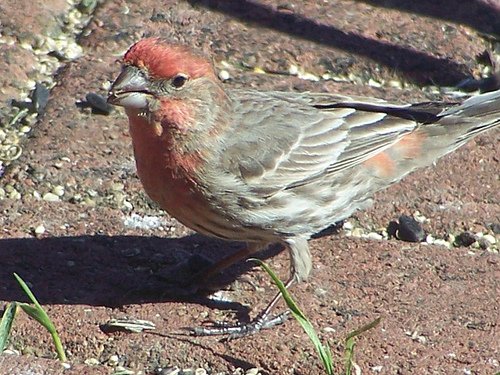}}};
    \end{scope}
    \node[blue] at (-6.2,1.8) {Train feature extractor};
    \node at (-6.2,1.0) {$\mathbf{x} \mapsto f_\varphi(\mathbf{x})\in{\left({\mathbb{R}^+}\right)}^d$};
    
    \node [input, name=input] at (-3, 5.25) {};
    \node [right of=input] (extractor) {$f_\varphi$};
    \node [block, right of=extractor,node distance=3cm] (transform) {PEME};
    \draw [draw,->] (input) -- node[pos=0.4] {\small $S\cup Q\in\mathbf{D}_{novel}$} (extractor);
    \draw [->] (extractor) -- node {} (transform);
    
    \node [block, right of=transform, node distance=6.25cm, pin={[pinstyle]right:Initialized $\mathbf{w}_j$}] (sinkhorn) {Min-size Sinkhorn};
    \node [block, below of=sinkhorn, node distance=1.8cm] (center) {Weight update};
    \node [right of=center, node distance=2.5cm] (step) {$n_{steps}$};
    \node [block, below of=center, node distance=1.8cm, pin={[pinstyle]below:$\mathbf{f}_Q$}] (prediction) {Prediction};
    \node [output, right of=prediction, node distance=3cm] (output) {};
     
    \draw [->] (transform) -- node[pos=0.5, name=f] {$\mathbf{f}_S\cup\mathbf{f}_Q$} (sinkhorn);
    \draw [->] (sinkhorn) -- node[name=m] {$\mathbf{P}$} (center);
    \draw [->] (center) -- node[name=c] {$\mathbf{w}_j$} (prediction);
    \draw [->] (prediction) -- node[name=o] {Accuracy} (output);
    \path
    (c) edge[bend right] (step);
    \path[->,>=stealth']
    (step) edge[bend right=25] (sinkhorn);
    
    \node at (2,3.5) (skew) {\fcolorbox{black}{white}{\includegraphics[width=0.25\linewidth,height=1.6cm]{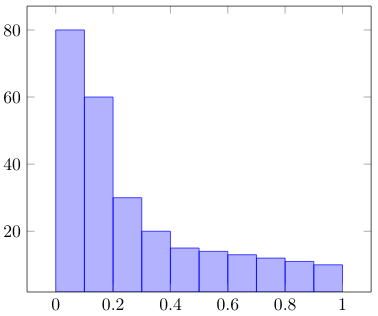}}};
    \node[font=\small] at (3,4) {$h_j(k)$};
    \node at (2,1) (gaussian) {\fcolorbox{black}{white}{\includegraphics[width=0.25\linewidth,height=1.6cm]{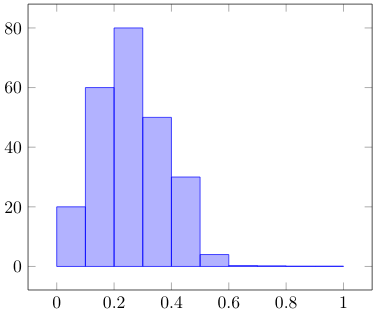}}};
    \node[font=\small] at (3,1.5) {$\tilde{h}_j(k)$};
    \draw [->] (skew) -- node[name=pt] {PEME} (gaussian);
    
    \end{tikzpicture}
    \caption{Illustration of the proposed method. First we train a feature extractor $f_\varphi$ using $\mathbf{D}_{base}$ that has a large number of labelled data. Then we extract feature vectors of all the inputs (support set $S$ and query set $Q$) in $\mathbf{D}_{novel}$ (the considered few-shot dataset). We preprocess them with proposed PEME, which contains power transform that has the effect of mapping a skewed feature distribution into a gaussian-like distribution ($h_j(k)$ denotes the histogram of feature $k$ in class $j$). The result feature vectors are denoted by $\mathbf{f}_S\cup\mathbf{f}_Q$. In the case of transductive learning, we introduce another step called Boosted Min-size Sinkhorn (BMS), where we perform a modified Sinkhorn algorithm with class weight parameters $\mathbf{w}_j$ initialized on labelled feature vectors $\mathbf{f}_S$ to obtain the class allocation matrix $\mathbf{P} $ for the inputs, and we update the weight parameters for the next iteration. After $n_{steps}$ we evaluate the accuracy on $\mathbf{f}_Q$.}
    \label{fig:illustration}
\end{figure*}
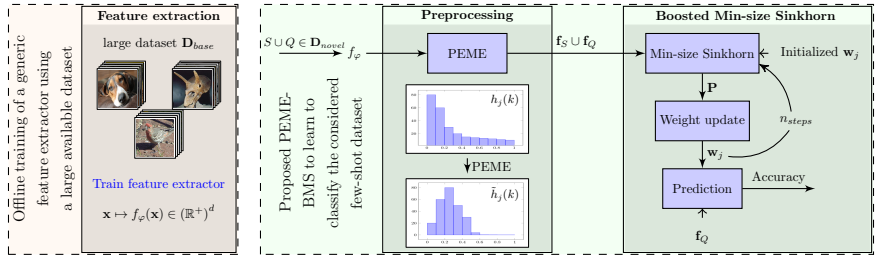

\section{Related work}
\label{related work}

A large volume of works in few-shot classification is based on meta learning~\cite{thrun2012learning} methods, where the training data is transformed into few-shot learning episodes to better fit in the context of few examples. In this branch, optimization based methods~\cite{thrun2012learning, finn2017model, DBLP:conf/iclr/RaviL17, DBLP:journals/corr/LiZCL17, DBLP:conf/iclr/AntoniouES19, DBLP:conf/iclr/BertinettoHTV19} train a well-initialized optimizer so that it quickly adapts to unseen classes with a few epochs of training. Other works~\cite{zhang2019few,chen2019image} apply data augmentation techniques to artificially increase the size of the training data in order for the model to generalize better to unseen data.

In the past few years, there have been a growing interest in transfer-based methods. The main idea consists in training feature extractors able to efficiently segregate novel classes it never saw before. For example, in~\cite{DBLP:conf/iclr/ChenLKWH19} the authors train the backbone with a distance-based classifier~\cite{mensink2012metric} that takes into account the inter-class distance. In~\cite{mangla2020charting}, the authors utilize self-supervised learning techniques~\cite{chapelle2009semi} to co-train an extra rotation classifier for the output features, improving the accuracy in few-shot settings. Aside from approaches focused on training a more robust model, other approaches are built on top of a pre-trained feature extractor (backbone). For instance, in~\cite{DBLP:journals/corr/abs-1911-04623} the authors implement a nearest class mean classifier to associate an input with a class whose centroid is the closest in terms of the $\ell_2$ distance. In~\cite{lichtenstein2020tafssl} an iterative approach is used to adjust the class prototypes. In~\cite{hu2021graph} the authors build a graph neural network to gather the feature information from similar samples. Generally, transfer-based techniques often reach the best performance on standardized benchmarks.

Although many works involve feature extraction, few have explored the features in terms of their distribution~\cite{8462273, DBLP:conf/iclr/YangLX21, mangla2020charting}. Often, assumptions are made that the features in a class align to a certain distribution, even though these assumptions are seldom experimentally discussed. In our work, we analyze the impact of the features distributions and how they can be transformed for better processing and accuracy. We also introduce a new algorithm to improve the quality of the association between input features and corresponding classes in typical few-shot settings.

\textbf{Contributions.} Let us highlight the main contributions of this work. (1) We propose to preprocess the raw extracted features in order to make them more aligned with Gaussian assumptions. Namely we introduce transforms of the features so that they become less skewed. (2) We use a Wasserstein-based method to better align the distribution of features with that of the considered classes. (3) We show that the proposed method can bring large increase in accuracy with a variety of feature extractors and datasets, leading to state-of-the-art results in the considered benchmarks. This work is an extended version of~\cite{hu2021leveraging}, with the main difference that here we consider the broader case where we do not know the proportion of samples belonging to each considered class in the case of transductive few-shot, leading to a new algorithm called Boosted Min-size Sinkhorn. We also propose more efficient preprocessing steps, leading to overall better performance in both inductive and transductive settings. Finally, we introduce the use of Logistic Regression in our methodology instead of a simple Nearest Class Mean classifier.

\section{Methodology}
\label{methodology}

In this section we introduce the problem statement. We also discuss the various steps of the proposed method, including training the feature extractors, preprocessing the feature representations, and classifying them. Note that we made the code of our method available at~\url{https://github.com/yhu01/BMS}.

\subsection{Problem statement}

We consider a typical few-shot learning problem. Namely, we are given a \emph{base} dataset $\mathbf{D}_{base}$ and a \emph{novel} dataset $\mathbf{D}_{novel}$ such that $\mathbf{D}_{base}\cap\mathbf{D}_{novel}=\emptyset$. $\mathbf{D}_{base}$ contains a large number of labelled examples from $K$ different classes and can be used to train a generic feature extractor. $\mathbf{D}_{novel}$, also referred to as a task or episode in other works, contains a small number of labelled examples (support set $\mathbf{S}$), along with some unlabelled ones (query set $\mathbf{Q}$), all from $n$ \emph{new} classes that are distinct from the $K$ classes in $\mathbf{D}_{base}$. Our goal is to predict the classes of unlabelled examples in the query set. The following parameters are of particular importance to define such a few-shot problem: the number of classes in the novel dataset $n$ (called $n$-way), the number of labelled samples per class $s$ (called $s$-shot) and the number of unlabelled samples per class $q$. Therefore, the novel dataset contains a total of $l+u$ samples, where $l=ns$ are labelled, and $u=nq$ are unlabelled.
In the case of inductive few-shot, the prediction is performed independently on each one of the query samples. In the case of transductive few-shot~\cite{DBLP:conf/iclr/LiuLPKYHY19, lichtenstein2020tafssl}, the prediction is performed considering all unlabelled samples together. Contrary to our previous work~\cite{hu2021leveraging}, we do not consider knowing the proportion of samples in each class in the case of transductive few-shot.

\subsection{Feature extraction}

The first step is to train a neural network backbone model using only the base dataset. In this work we consider multiple backbones, with various training procedures. 
Once the considered backbone is trained, we obtain robust embeddings that should generalize well to novel classes. We denote by $f_\varphi$ the backbone function, obtained by extracting the output of the penultimate layer from the considered architecture, with $\varphi$ being the trained architecture parameters. Thus considering an input vector $\mathbf{x}$, $ f_\varphi(\mathbf{x})$ is a feature vector with $d$ dimensions that can be thought of as a simpler-to-manipulate representation of $\mathbf{x}$.
Note that importantly, in all backbone architectures used in the experiments of this work, the penultimate layers are obtained by applying a ReLU function, so that all feature components coming out of $f_\varphi$ are nonnegative.

\subsection{Feature preprocessing}

As mentioned in Section~\ref{related work}, many works hypothesize, explicitly or not, that the features from the same class are aligned with a specific distribution (often Gaussian-like). But this aspect is rarely experimentally verified. In fact, it is very likely that features obtained using the backbone architecture are not Gaussian. Indeed, usually the features are obtained after applying a ReLU function~\cite{DBLP:journals/corr/abs-1803-08375}, and exhibit a positive and yet skewed distribution mostly concentrated around 0 (more details can be found in the next section). 


Multiple works in the domain~\cite{DBLP:journals/corr/abs-1911-04623, lichtenstein2020tafssl} discuss the different statistical methods (e.g. batch normalization) to better fit the features into a model. Although these methods may have provable assets for some distributions, they could worsen the process if applied to an unexpected input distribution. This is why we propose to preprocess the obtained raw feature vectors so that they better align with typical distribution assumptions in the field. Denote $ f_\varphi(\mathbf{x})=[f^{1}_\varphi(\mathbf{x}), ..., f^{h}_\varphi(\mathbf{x}), ..., f^{d}_\varphi(\mathbf{x})]\in{\left({\mathbb{R}^+}\right)}^d, \mathbf{x}\in \mathbf{D}_{novel}$ as the obtained features on $\mathbf{D}_{novel}$, and $f^{h}_\varphi(\mathbf{x}), 1\leq h\leq d$ denotes its value in the h\textsuperscript{th} position. The preprocessing methods applied in our proposed algorithms are as follows:  

\textbf{Euclidean normalization.} Also known as L2-normalization that is widely used in many related works~\cite{DBLP:journals/corr/abs-1911-04623, DBLP:conf/iclr/YangLX21, hu2021graph}, this step scales the features to the same area so that large variance feature vectors do not predominate the others. Euclidean normalization can be given by:
\begin{equation}
f_\varphi(\mathbf{x}) \leftarrow \frac{f_\varphi(\mathbf{x})}{\|f_\varphi(\mathbf{x})\|_2}
\label{eq:preprocessing_norm}
\end{equation}

\textbf{Power transform.} Power transform method~\cite{tukey1977exploratory,hu2021leveraging} simply consists of taking the power of each feature vector coordinate. The formula is given by:
\begin{equation}
f^{h}_\varphi(\mathbf{x}) \leftarrow (f^{h}_\varphi(\mathbf{x})+\epsilon)^{\beta}, \hspace{0.2cm} \beta \neq 0 
\label{eq:preprocessing_pt}
\end{equation}
where $\epsilon = 1e-6$ is used to make sure that $f_\varphi(\mathbf{x})+\epsilon$ is strictly positive in every position, and $\beta$ is a hyper-parameter. The rationale of the preprocessing above is that power transform, often used in combination with euclidean normalization, has the functionality of reducing the skew of a distribution and mapping it to a close-to-gaussian distribution, adjusted by $\beta$. After experiments, we found that $\beta=0.5$ gives the most consistent results for our considered experiments, which corresponds to a square-root function that has a wide range of usage on features~\cite{cinbis2015approximate}. We will analyse this ability and the effect of power transform in more details in Section~\ref{experiments}. Note that power transform can only be applied if considered feature vectors contain nonnegative entries, which will always be the case in the remaining of this work.

\textbf{Mean subtraction.} With mean subtraction, each sample is translated using $\mathbf{\mathbf{m}}\in{\left({\mathbb{R}^+}\right)}^d$, the projection center. This is often used in combination with euclidean normalization in order to reduce the task bias and better align the feature distributions~\cite{lichtenstein2020tafssl}. The formula is given by: 
\begin{equation}
f_\varphi(\mathbf{x}) \leftarrow f_\varphi(\mathbf{x}) - \mathbf{\mathbf{m}}
\label{eq:preprocessing_submean}
\end{equation}
The projection center is often computed as the mean values of feature vectors related to the problem~\cite{DBLP:journals/corr/abs-1911-04623, lichtenstein2020tafssl}. In this paper we compute it either as the mean feature vector of the base dataset (denoted as $\mathrm{M_b}$) or the mean vector of the novel dataset (denoted as $\mathrm{M_n}$), depending on the few-shot settings. Of course, in both of these cases, the rationale is to consider a proxy to what would be the exact mean value of feature vectors on the considered task.

In our proposed method we deploy these preprocessing steps in the following order: Power transform (P) on the raw features, followed by an Euclidean normalization (E). Then we perform Mean subtraction (M) followed by another Euclidean normalization at the end. For simplicity we denote PEME as our proposed preprocessing order, in which M can be either $\mathrm{M_b}$ or $\mathrm{M_n}$ as mentioned above. In our experiments, we found that using $\mathrm{M_b}$ in the case of inductive few-shot learning and $\mathrm{M_n}$ in the case of transductive few-shot learning consistently led to the most competitive results. More details on why we used this methodology are available in the experiment section.


When facing an inductive problem, a simple classifier such as a Nearest-Class-Mean classifier (NCM) can be used directly after this preprocessing step. The resulting methodology is denoted PE$\mathrm{M_b}$E-NCM. But in the case of transductive settings, we also introduce an iterative procedure, denoted BMS for Boosted Min-size Sinkhorn, meant to leverage the joint distribution of unlabelled samples. The resulting methodology is denoted PE$\mathrm{M_n}$E-BMS. The details of the BMS procedure are presented thereafter.


\subsection{Boosted Min-size Sinkhorn}

In the case of transductive few-shot, we introduce a method that consists in iteratively refining estimates for the probability each unlabelled sample belong to any of the considered classes. This method is largely based on the one we introduced in~\cite{hu2021leveraging}, except it does not require priors about samples distribution in each of the considered class. Denote $i\in[1,...,l+u]$ as the sample index in $\mathbf{D}_{novel}$ and $j\in[1,...,n]$ as the class index, the goal is to maximize the following log post-posterior function: 
\begin{equation}
\begin{split}
    L(\theta)&= \sum_i \log P(l(\mathbf{x}_i)=j | \mathbf{x}_i ; \theta) \\
    = & \sum_i \log \frac{P(\mathbf{x}_i, l(\mathbf{x}_i)=j ; \theta) }{P(\mathbf{x}_i; \theta)} \\
    \propto & \sum_i \log \frac{P(\mathbf{x}_i | l(\mathbf{x}_i)=j ; \theta)}{ P(\mathbf{x}_i; \theta)},
\end{split}
\label{eq:MAP}
\end{equation}
here $l(\mathbf{x}_i)$ denotes the class label for sample $\mathbf{x}_i\in\mathbf{Q}	\cup\mathbf{S}$, $P(\mathbf{x}_i; \theta)$ denotes the marginal probability, and $\theta$ represents the model parameters to estimate. Assuming a gaussian distribution on the input features for each class, here we define $\theta=\mathbf{w}_j, \forall j$ where $\mathbf{w}_j\in\mathbb{R}^d$ stand for the weight parameters for class $j$. We observe that Eq.~\ref{eq:MAP} can be related to the cost function utilized in Optimal Transport~\cite{villani2008optimal}, which is often considered to solve classification problems, with constrains on the sample distribution over classes. To that end, a well-known  Sinkhorn~\cite{cuturi2013sinkhorn} mapping method is proposed. The algorithm aims at computing a class allocation matrix among novel class data for a minimum Wasserstein distance. Namely, an allocation matrix $\mathbf{P}\in\mathbb{R}_{+}^{(l+u) \times n}$ is defined where $\mathbf{P}[i,j]$ denotes the assigned portion for sample $i$ to class $j$, and it is computed as follows:
\begin{equation}
\begin{split}
\mathbf{P}&= Sinkhorn(\mathbf{C}, \mathbf{p}, \mathbf{q}, \lambda) \\
&= \underset{\mathbf{\tilde{P}}\in\mathbb{U}(\mathbf{p},\mathbf{q})}{\mathrm{argmin}}\, \sum_{ij}\mathbf{\tilde{P}}[i,j]\mathbf{C}[i,j]  + \lambda H(\mathbf{\tilde{P}}),
\end{split}
\label{eq:P}
\end{equation}
where $\mathbb{U}(\mathbf{p},\mathbf{q})\in\mathbb{R}_{+}^{(l+u) \times n}$ is a set of positive matrices for which the rows sum to $\mathbf{p}$ and the columns sum to $\mathbf{q}$, $\mathbf{p}$ denotes the distribution of the amount that each sample uses for class allocation, and $\mathbf{q}$ denotes the distribution of the amount of samples allocated to each class. Therefore, $\mathbb{U}(\mathbf{p},\mathbf{q})$ contains all the possible ways of allocation. In the same equation, $\mathbf{C}$ can be viewed as a cost matrix that is of the same size as $\mathbf{P}$, each element in $\mathbf{C}$ indicates the cost of its corresponding position in $\mathbf{P}$. We will define the particular formula of the cost function for each position $\mathbf{C}[i,j], \forall i,j$ in details later on in the section. As for the second term on the right of ~\ref{eq:P}, it stands for the entropy of $\mathbf{\tilde{P}}$: $H(\mathbf{\tilde{P}})=-\sum_{ij}\mathbf{\tilde{P}}[i,j]\log \mathbf{\tilde{P}}[i,j]$, regularized by a hyper-parameter $\lambda$. Increasing $\lambda$ would force the entropy to become smaller, so that the mapping is less diluted. This term also makes the objective function strictly convex~\cite{cuturi2013sinkhorn, solomon2015convolutional} and thus a practical and effective computation. 
From lemma 2 in~\cite{cuturi2013sinkhorn}, the result of Sinkhorn allocation has the typical form $\mathbf{P} = \text{diag}(\mathbf{u}) \cdot \exp(-\mathbf{C}/\lambda) \cdot \text{diag}(\mathbf{v})$. It is worth noting that here we assume a soft class allocation, meaning that each sample can be ``sliced'' into different classes. We will present our proposed method in details in the next paragraphs.

Given all that are presented above, in this paper we propose an Expectation–Maximization (\emph{EM})~\cite{dempster1977maximum} based method which alternates between updating the allocation matrix $\mathbf{P}$ and estimating the parameter $\theta$ of the designed model, in order to minimize Eq.~\ref{eq:P} and maximize Eq.~\ref{eq:MAP}. For a starter, we define a weight matrix $\mathbf{W}$ with $n$ columns (i.e one per class) and $d$ rows (i.e one per dimension of feature vectors), for column $j$ in $\mathbf{W}$ we denote it as the weight parameters $\mathbf{w}_j\in\mathbb{R}^d$ for class $j$ in correspondence with Eq.~\ref{eq:MAP}. And it is initialized as follows:
\begin{equation}
    \mathbf{w}_j = \mathbf{W}[:, j] = \mathbf{c}_j / \|\mathbf{c}_j\|_2, 
\label{eq:init_W_1}
\end{equation}
where 
\begin{equation}
\mathbf{c}_j = \frac{1}{s} \sum_{\mathbf{x}\in \mathbf{S}, \ell(\mathbf{x}) = j}{f_\varphi(\mathbf{x})}.
\label{eq:init_c}
\end{equation}
We can see that $\mathbf{W}$ contains the average of feature vectors in the support set for each class, followed by a L2-normalization on each column so that $\|\mathbf{w}_j\|_2=1, \forall j$. 

Then, we iterate multiple steps that we describe thereafter.

\textbf{a. Computing costs}

As previously stated, the proposed algorithm is an \emph{EM}-like one that iterately updates model parameters for optimal estimates. Therefore, this step along with Min-size Sinkhorn presented in the next step, is considered as the \emph{E}-step of our proposed method. The goal is to find membership probabilities for the input samples, namely, we compute $\mathbf{P}$ that minimizes Eq.~\ref{eq:P}.

Here we assume gaussian distributions, features in each class have the same variance and are independent from one another (covariance matrix $\mathbf{\Sigma}=\mathbf{I}\sigma^2$). We observe that, ignoring the marginal probability, Eq.~\ref{eq:MAP} can be boiled down to negative L2 distances between extracted samples $f_\varphi(\mathbf{x}_i), \forall i$ and $\mathbf{w}_j, \forall j$, which is initialized in Eq.~\ref{eq:init_W_1} in our proposed method. Therefore, based on the fact that $\mathbf{w}_j$ and $f_\varphi(\mathbf{x}_i)$ are both normalized to be unit length vectors ($f_\varphi(\mathbf{x}_i)$ being preprocessed using PEME introduced in the previous section), here we define the cost between sample $i$ and class $j$ to be the following equation:
\begin{equation}
\begin{split}
    \mathbf{C}[i,j] \propto (f_\varphi(\mathbf{x}_i)-\mathbf{w}_{j})^2 \\
    = 1-\mathbf{w}_{j}^{T} f_\varphi(\mathbf{x}_i),
\end{split}
\label{eq:C}
\end{equation}
which corresponds to the cosine distance.


\textbf{b. Min-size Sinkhorn}

In~\cite{hu2021leveraging}, we proposed a Wasserstein distance based method in which the Sinkhorn algorithm is applied at each iteration so that the class prototypes are updated iteratively in order to find their best estimates. Although the method showed promising results, it is established on the condition that the distribution of the query set is known, e.g. a uniform distribution among classes on the query set. This is not ideal given the fact that any priors about $\mathbf{Q}$ should be supposedly kept unknown when applying a method. The methodology introduced in this paper can be seen as a generalization of that introduced in~\cite{hu2021leveraging} that does not require priors about $\mathbf{Q}$.

In the classical settings, Sinkhorn algorithm aims at finding the optimal matrix $\mathbf{P}$, given the cost matrix $\mathbf{C}$ and regulation parameter $\lambda$ presented in Eq.~\ref{eq:MAP}). Typically it initiates $\mathbf{P}$ from a softmax operation over the rows in $\mathbf{C}$, then it iterates between normalizing columns and rows of $\mathbf{P}$, until the resulting matrix becomes close-to doubly stochastic according to $\mathbf{p}$ and $\mathbf{q}$. However, in our case we do not know the distribution of samples over classes. To address this, we firstly introduce the parameter $k$, initialized so that $k\leftarrow s$, meant to track an estimate of the cardinal of the class containing the least number of samples in the considered task. Then we propose the following modification to be applied to the matrix $\mathbf{P}$ once initialized: we normalize each row as in the classical case, but only normalize the columns of $\mathbf{P}$ for which the sum is less than the previously computed min-size $k$~\cite{lichtenstein2020tafssl}. This ensures at least $k$ elements allocated for each class, but not exactly $k$ samples as in the balanced case. 

The principle of this modified Sinkhorn solution is presented in Algorithm~\ref{alg:boostingSinkhorn}.

\begin{algorithm}[tb]
   \caption{Min-size Sinkhorn}
   \label{alg:boostingSinkhorn}
\begin{algorithmic}
   \STATE {\bfseries Inputs:} {$\mathbf{C}, \mathbf{p}= \mathbf{1}_{l+u}, \mathbf{q}=k\mathbf{1}_{n}$, $\lambda$}
   \STATE {\bfseries Initializations:} {$\mathbf{P}= Softmax{(-\lambda\mathbf{C}})$}
   \FOR{$iter=1$ {\bfseries to} $50$}
   \STATE $\mathbf{P}[i,:] \leftarrow \mathbf{p}[i]\cdot\frac{\mathbf{P}[i,:]}{\sum_j{\mathbf{P}[i,j]}}, \forall i$
   \STATE $\mathbf{P}[:,j] \leftarrow \mathbf{q}[j]\cdot\frac{\mathbf{P}[:,j]}{\sum_i{\mathbf{P}[i,j]}} \text{ if } \sum_i{\mathbf{P}[i,j]} < \mathbf{q}[j], \forall j$
   \ENDFOR
   
\STATE {\bfseries return} $\mathbf{P}$
\end{algorithmic}
\end{algorithm}

\textbf{c. Updating weights}

This step is considered as the \emph{M}-step of the proposed algorithm, in which we use a variant of the Logistic Regression algorithm in order to find the model parameter $\theta$ in the form of weight parameters $\mathbf{w}_j$ for each class. Note that $\mathbf{w}_j$, if normalized, is equivalent to the prototype for class $j$ in this case. Given the fact that in Eq.~\ref{eq:MAP} we also take into account the marginal probability, which can be further broken down as:
\begin{equation}
P(\mathbf{x}_i;\theta) = \sum_j P(\mathbf{x}_i | l(\mathbf{x}_i)=j;\theta)P(l(\mathbf{x}_i)=j),
\label{eq:margin}
\end{equation}
we observe that Eq.~\ref{eq:MAP} corresponds to applying a softmax function on the negative logits computed through a L2-distance function between samples and class prototypes (normalized). This fits the formulation of a linear hypothesis between $f_\varphi(\mathbf{x}_i)$ and $\mathbf{w}_j$ for logit calculations, hence the rationale for utilizing Logistic Regression in our proposed method.    

The procedure of this step is as follows: now that we have a polished allocation matrix $\mathbf{P}$, we firstly initialize the weights $\mathbf{w}_j$ as follows:
\begin{equation}
\mathbf{w}_j \leftarrow \mathbf{u}_j / \|\mathbf{u}_j\|_2,
\label{eq:init_W_2}
\end{equation}
where 
\begin{equation}
\mathbf{u}_j \leftarrow \sum_{i} \mathbf{P}[i,j]f_\varphi(\mathbf{x}_i) / \sum_{i}\mathbf{P}[i,j].
\label{eq:center_update}
\end{equation}
We can see that elements in $\mathbf{P}$ are used as coefficients for feature vectors to linearly adjust the class prototypes~\cite{hu2021leveraging}. Similar to Eq.~\ref{eq:init_W_1}, here $\mathbf{w}_j$ is the normalized newly-computed class prototype that is a vector of length $1$. 

Next we further adjust weights by applying a logistic regression, the optimization is performed by minimizing the following loss:
\begin{equation}
    \frac{1}{l+u}\cdot\sum_i\sum_j -log(\frac{\exp{(\mathbf{S}[i,j])}}{\sum_{\gamma=1}^{n} \exp{(\mathbf{S}[i,\gamma]})})\cdot\mathbf{P}[i,j],
\label{eq:LR}
\end{equation}
where $\mathbf{S}\in\mathbb{R}^{(l+u)\times n}$ contains the logits, each element is computed as:
\begin{equation}
\mathbf{S}[i,j] = \kappa\cdot\frac{\mathbf{w}_{j}^{T} f_\varphi(\mathbf{x}_i)}{\|\mathbf{w}_j\|_2}.
\label{eq:logits}
\end{equation}
Note that $\kappa$ is a scaling parameter, it can also be seen as a temperature parameter that adjusts the confidence metric to be associated to each sample. And it is learnt jointly with $\mathbf{W}$.

The deployed Logistic Regression comes with hyperparameters on its own. In our experiments, we use an SGD optimizer with a gradient step of $0.1$ and $0.8$ as the momentum parameter, and we train over $e$ epochs. Here we point out that $e\geq 0$ is considered an influential hyperparameter in our proposed algorithm, $e=0$ indicates a simple update of $\mathbf{W}$ as the normalized adjusted class prototypes~(Eq.~\ref{eq:init_W_2}) computed from $\mathbf{P}$ in Eq.~\ref{eq:center_update}, without further adjustment of logistic regression. And also note that when $e>0$ we project columns of $\mathbf{W}$ to the unit hypersphere at the end of each epoch. 

\textbf{d. Estimating the class minimum size}

We can now refine our estimate for the min-size $k$ for the next iteration. To this end, we firstly compute the predicted label of each sample as follows: 
\begin{equation}
\hat{\ell}(\mathbf{x}_i)=\arg\max_j(\mathbf{P}[i,j]),
\label{eq:predictions}
\end{equation} 
which can be seen as the current (temporary) class prediction.

Then, we compute:
\begin{equation}
k = \min_{j}{\{k_j\}},
\label{eq:k}
\end{equation} 
where $k_j = \#\{i, \hat{\ell}(\mathbf{x}_i) = j\}$, $\#\{\cdot\}$ representing the cardinal of a set.

\textbf{Summary of the proposed method.} All steps of the proposed method are summarized in Algorithm~\ref{alg:CSC}. In our experiments, we also report the results obtained when using a prior about $\mathbf{Q}$ as in~\cite{hu2021leveraging}. In this case, $k$ does not have to be estimated throughout the iterations and can be replaced with the actual exact targets for the Sinkhorn. We denote this prior-dependent version PE$\mathrm{M_n}$E-BMS$^*$ (with an added $*$).

\begin{algorithm}[tb]
   \caption{Boosted Min-size Sinkhorn~(BMS)}
   \label{alg:CSC}
\begin{algorithmic}
   \STATE {\bfseries Parameters:} {$\lambda, e$}
   \STATE {\bfseries Inputs:} {Preprocessed $f_\varphi(\mathbf{x})$, $\forall\mathbf{x}\in\mathbf{D}_{novel}=\mathbf{Q}\cup\mathbf{S}$}
   \STATE {\bfseries Initializations:} $\mathbf{W}$ as normalized mean vectors over the support set for each class (Eq.~\ref{eq:init_W_1}); Min-size $k \leftarrow s$.
   \FOR{$iter=1$ {\bfseries to} $20$}
   \STATE Compute cost matrix $\mathbf{C}$ using $\mathbf{W}$ (Eq.~\ref{eq:C}). \# $E$-step
   \STATE Apply Min-size Sinkhorn to compute $\mathbf{P}$ (Algorithm~\ref{alg:boostingSinkhorn}). \# $E$-step
   \STATE Update weights $\mathbf{W}$ using $\mathbf{P}$ with logistic regression (Eq.~\ref{eq:init_W_2}-\ref{eq:logits}). \# $M$-step
   \STATE Estimate class predictions $\hat{\ell}$ and min-size $k$ using $\mathbf{P}$ (Eq.~\ref{eq:predictions}-\ref{eq:k}).
   \ENDFOR
\STATE {\bfseries return} $\hat{\ell}$
\end{algorithmic}
\end{algorithm}

\section{Experiments}
\label{experiments}
\subsection{Datasets}

We evaluate the performance of the proposed method using standardized few-shot classification datasets: miniImageNet~\cite{vinyals2016matching}, tieredImageNet~\cite{DBLP:conf/iclr/RenTRSSTLZ18}, CUB~\cite{WahCUB_200_2011} and CIFAR-FS~\cite{DBLP:conf/iclr/BertinettoHTV19}. The \textbf{miniImageNet} dataset contains 100 classes randomly chosen from ILSVRC-
2012~\cite{russakovsky2015imagenet} and 600 images of size $84\times84$ pixels per class. It is split into 64 base classes, 16 validation classes and 20 novel classes. The \textbf{tieredImageNet} dataset is another subset of ImageNet, it consists of 34 high-level categories with 608 classes in total. These categories are split into 20 meta-training superclasses, 6 meta-validation superclasses and 8 meta-test superclasses, which corresponds to 351 base classes, 97 validation classes and 160 novel classes respectively. The \textbf{CUB} dataset contains 200 classes of birds and has 11,788 images of size $84\times84$ pixels in total, it is split into 100 base classes, 50 validation classes and 50 novel classes. The \textbf{CIFAR-FS} dataset has 100 classes, each class contains 600 images of size $32\times32$ pixels. The splits of this dataset are the same as those in miniImageNet.

\subsection{Implementation details}

In order to stress the genericity of our proposed method with regards to the chosen backbone architecture and training strategy, we perform experiments using \textbf{WRN}~\cite{DBLP:conf/bmvc/ZagoruykoK16}, \textbf{ResNet18} and \textbf{ResNet12}~\cite{he2016deep}, along with some other pretrained backbones (e.g. \textbf{DenseNet}~\cite{huang2017densely, DBLP:journals/corr/abs-1911-04623}). For each dataset we train the feature extractor with base classes and test the performance using novel classes. Therefore, for each test run, $n$ classes are drawn uniformly at random among novel classes. Among these $n$ classes, $s$ labelled examples and $q$ unlabelled examples per class are uniformly drawn at random to form $\mathbf{D}_{novel}$. The WRN and ResNet are trained following~\cite{mangla2020charting}. In the inductive setting, we use our proposed preprocessing steps PE$\mathrm{M_b}$E followed by a basic Nearest Class Mean (NCM) classifier. In the transductive setting, the preprocessing steps are denoted as PE$\mathrm{M_n}$E in that we use the mean vector of novel dataset for mean subtraction, followed by BMS or BMS$^*$ depending on whether we have prior knowledge on the distribution of query set $\mathbf{Q}$ among classes. Note that we perform a QR decomposition on preprocessed features in order to speed up the computation for the classifier that follows. All our experiments are performed using $n=5, q=15$, $s=1$ or $5$. We run 10,000 random draws to obtain mean accuracy score and indicate confidence scores ($95\%$) when relevant. For our proposed PE$\mathrm{M_n}$E-BMS, we train $e=0$ epoch in the case of 1-shot and $e=40$ epochs in the case of 5-shot. As for PE$\mathrm{M_n}$E-BMS$^*$ we set $e=20$ for 1-shot and $e=40$ for 5-shot. As for the regularization parameter $\lambda$ in Eq.~\ref{eq:P}, it is fixed to $8.5$ for all settings. Impact of these hyperparameters is detailed in the next sections.

\subsection{Comparison with state-of-the-art methods}
\textbf{Performance on standardized benchmarks.} In the first experiment, we conduct our proposed method on different benchmarks and compare the performance with other state-of-the-art solutions. The results are presented in Table~\ref{tab:results1} and \ref{tab:results2}, we observe that our method reaches the state-of-the-art performance in both inductive and transductive settings on all the few-shot classification benchmarks. Particularly, the proposed PE$\mathrm{M_n}$E-BMS$^*$ brings important gains in both 1-shot and 5-shot settings, and the prior-independent PE$\mathrm{M_n}$E-BMS also obtains competitive results on 5-shot. Note that for tieredImageNet we implement our method based on a pre-trained DenseNet121 backbone following the procedure described in~\cite{DBLP:journals/corr/abs-1911-04623}. From these experiments we conclude that the proposed method can bring an increase of accuracy with a variety of backbones and datasets, leading to state-of-the-art performance. In terms of execution time, we measured an average of $0.004s$ per run.

\begin{table*}[h!]
    \caption{1-shot and 5-shot accuracy of state-of-the-art methods in the literature on miniImageNet and tieredImageNet, compared with the proposed solution.}
    \centering
    \scalebox{0.7}{
    \begin{tabular}{c|l|l|l|l}
         \toprule
         &     &     & \multicolumn{2}{c}{\textbf{miniImageNet}} \\
         Setting & Method & Backbone & 1-shot & 5-shot \\
         \midrule
         \multirow{7}{*}{Inductive}
         &Matching Networks~\cite{vinyals2016matching} & WRN & $64.03\pm0.20\%$ & $76.32\pm0.16\%$\\ 
         &SimpleShot~\cite{DBLP:journals/corr/abs-1911-04623} & DenseNet121 & $64.29\pm0.20\%$ & $81.50\pm0.14\%$\\
         &S2M2\_R~\cite{mangla2020charting} & WRN & $64.93\pm0.18\%$ & $83.18\pm0.11\%$\\
         &PT+NCM~\cite{hu2021leveraging} & WRN & $65.35\pm0.20\%$ & $83.87\pm0.13\%$\\
         &DeepEMD\cite{zhang2020deepemd} & ResNet12 & $65.91\pm0.82\%$ & $82.41\pm0.56\%$\\
         &FEAT\cite{ye2020few} & ResNet12 & $66.78\pm0.20\%$ & $82.05\pm0.14\%$\\
         &PE$\mathrm{M_b}$E-NCM (ours) & WRN & $\mathbf{68.43\pm0.20}\%$ & $\mathbf{84.67\pm0.13}\%$ \\
         \midrule
         \multirow{10}{*}{Transductive}
         &BD-CSPN~\cite{liu2020prototype} & WRN & $70.31\pm0.93\%$ & $81.89\pm0.60\%$\\
         &LaplacianShot~\cite{ziko2020laplacian} & DenseNet121 & $75.57\pm0.19\%$ & $87.72\pm0.13\%$\\
         &Transfer+SGC~\cite{hu2021graph} & WRN & $76.47\pm0.23\%$ & $85.23\pm0.13\%$\\
         &TAFSSL~\cite{lichtenstein2020tafssl} & DenseNet121 & $77.06\pm0.26\%$ & $84.99\pm0.14\%$\\
         &TIM-GD~\cite{DBLP:journals/corr/abs-2008-11297} & WRN & $77.80\%$ & $87.40\%$\\
         &MCT~\cite{DBLP:journals/corr/abs-2002-12017} & ResNet12 & $78.55\pm0.86\%$ & $86.03\pm0.42\%$\\
         &EPNet~\cite{rodriguez2020embedding} & WRN & $79.22\pm0.92\%$ & $88.05\pm0.51\%$\\
         &PT+MAP~\cite{hu2021leveraging} & WRN & $82.92\pm0.26\%$ & $88.82\pm0.13\%$\\
         &PE$\mathrm{M_n}$E-BMS (ours) & WRN & $82.07\pm0.25\%$ & $89.51\pm0.13\%$ \\
         &PE$\mathrm{M_n}$E-BMS$^*$ (ours) & WRN & $\mathbf{83.35\pm0.25}\%$ & $\mathbf{89.53\pm0.13}\%$\\
         \bottomrule
         
         \toprule
         &  &          & \multicolumn{2}{c}{\textbf{tieredImageNet}} \\
         Setting & Method & Backbone & 1-shot & 5-shot \\       
         \midrule
         \multirow{8}{*}{Inductive}
         &ProtoNet~\cite{snell2017prototypical} & ConvNet4 & $53.31\pm0.89\%$ & $72.69\pm0.74\%$\\
         &LEO~\cite{DBLP:conf/iclr/RusuRSVPOH19} & WRN & $66.33\pm0.05\%$ & $81.44\pm0.09\%$\\
         &SimpleShot~\cite{DBLP:journals/corr/abs-1911-04623} & DenseNet121 & $71.32\pm0.22\%$ & $86.66\pm0.15\%$\\
         &PT+NCM~\cite{hu2021leveraging} & DenseNet121 & $69.96\pm0.22\%$ & $86.45\pm0.15\%$\\ 
         &FEAT\cite{ye2020few} & ResNet12 & $70.80\pm0.23\%$ & $84.79\pm0.16\%$\\
         &DeepEMD\cite{zhang2020deepemd} & ResNet12 & $71.16\pm0.87\%$ & $86.03\pm0.58\%$\\
         &RENet\cite{DBLP:journals/corr/abs-2108-09666} & ResNet12 & $71.61\pm0.51\%$ & $85.28\pm0.35\%$\\
         &PE$\mathrm{M_b}$E-NCM (ours) & DenseNet121 & $\mathbf{71.86\pm0.21\%}$ & $\mathbf{87.09\pm0.15}\%$ \\
         \midrule
         \multirow{8}{*}{Transductive}
         &BD-CSPN~\cite{liu2020prototype} & WRN & $78.74\pm0.95\%$ & $86.92\pm0.63\%$\\
         &LaplacianShot~\cite{ziko2020laplacian} & DenseNet121 & $80.30\pm0.22\%$ & $87.93\pm0.15\%$\\
         &MCT~\cite{DBLP:journals/corr/abs-2002-12017} & ResNet12 & $82.32\pm0.81\%$ & $87.36\pm0.50\%$\\
         &TIM-GD~\cite{DBLP:journals/corr/abs-2008-11297} & WRN & $82.10\%$ & $89.80\%$\\
         &TAFSSL~\cite{lichtenstein2020tafssl} & DenseNet121 & $84.29\pm0.25\%$ & $89.31\pm0.15\%$\\
         &PT+MAP~\cite{hu2021leveraging} & DenseNet121 & $85.75\pm0.26\%$ & $90.43\pm0.14\%$\\ 
         &PE$\mathrm{M_n}$E-BMS (ours) & DenseNet121 & $85.08\pm0.25\%$ & $91.08\pm0.14\%$\\
         &PE$\mathrm{M_n}$E-BMS$^*$ (ours) & DenseNet121 & $\mathbf{86.07\pm0.25}\%$ & $\mathbf{91.09\pm0.14}\%$\\
         \bottomrule
         
    \end{tabular}
    }
    \label{tab:results1}
\end{table*}

\textbf{Performance on cross-domain settings.} In this experiment we test our method in a cross-domain setting, where the backbone is trained with the base classes in miniImageNet but tested with the novel classes in CUB dataset. As shown in Table~\ref{tab:results_cross}, the proposed method gives the best accuracy both in the case of 1-shot and 5-shot, for both inductive and transductive settings.

\begin{table}[h!]
    \caption{1-shot and 5-shot accuracy of state-of-the-art methods on CUB and CIFAR-FS.}
    \centering
    \scalebox{0.70}{
    \begin{tabular}{c|l|l|l|l}
         \toprule
         &     &      & \multicolumn{2}{c}{\textbf{CUB}} \\
         Setting & Method & Backbone & 1-shot & 5-shot \\       
         \midrule
         \multirow{10}{*}{Inductive}
         &Baseline++~\cite{DBLP:conf/iclr/ChenLKWH19} & ResNet10 & $69.55\pm0.89\%$ & $85.17\pm0.50\%$\\
         &MAML~\cite{finn2017model} & ResNet10 & $70.32\pm0.99\%$ & $80.93\pm0.71\%$\\
         &ProtoNet~\cite{snell2017prototypical} & ResNet18 & $72.99\pm0.88\%$ & $86.64\pm0.51\%$\\
         &Matching Networks~\cite{vinyals2016matching} & ResNet18 & $73.49\pm0.89\%$ & $84.45\pm0.58\%$\\ 
         &FEAT\cite{ye2020few} & ResNet12 & $73.27\pm0.22\%$ & $85.77\pm0.14\%$\\
         &DeepEMD\cite{zhang2020deepemd} & ResNet12 & $75.65\pm0.83\%$ & $88.69\pm0.50\%$\\
         &RENet\cite{DBLP:journals/corr/abs-2108-09666} & ResNet12 & $79.49\pm0.44\%$ & $91.11\pm0.24\%$\\
         &S2M2\_R~\cite{mangla2020charting} & WRN & $80.68\pm0.81\%$ & $90.85\pm0.44\%$\\
         &PT+NCM~\cite{hu2021leveraging} & WRN & $80.57\pm0.20\%$ & $91.15\pm0.10\%$\\
         &PE$\mathrm{M_b}$E-NCM (ours) & WRN & $\mathbf{80.82\pm0.19}\%$ & $\mathbf{91.46\pm0.10}\%$\\
         \midrule
         \multirow{7}{*}{Transductive}
         &LaplacianShot~\cite{ziko2020laplacian} & ResNet18 & $80.96\%$ & $88.68\%$\\
         &TIM-GD~\cite{DBLP:journals/corr/abs-2008-11297} & ResNet18 & $82.20\%$ & $90.80\%$\\
         &BD-CSPN~\cite{liu2020prototype} & WRN & $87.45\%$ & $91.74\%$\\
         &Transfer+SGC~\cite{hu2021graph} & WRN & $88.35\pm0.19\%$ & $92.14\pm0.10\%$\\
         &PT+MAP~\cite{hu2021leveraging} & WRN & $91.55\pm0.19\%$ & $93.99\pm0.10\%$ \\
         &LST+MAP~\cite{DBLP:journals/corr/abs-2102-05176} & WRN & $91.68\pm0.19\%$ & $94.09\pm0.10\%$ \\
         &PE$\mathrm{M_n}$E-BMS (ours) & WRN & $91.01\pm0.19\%$ & $94.60\pm0.09\%$\\
         &PE$\mathrm{M_n}$E-BMS$^*$ (ours) & WRN & $\mathbf{91.91\pm0.18}\%$ & $\mathbf{94.62\pm0.09}\%$\\ 
         \bottomrule
         
         \toprule
         &     &     & \multicolumn{2}{c}{\textbf{CIFAR-FS}} \\
         Setting & Method & Backbone & 1-shot & 5-shot \\
         \midrule
         \multirow{7}{*}{Inductive}
         &ProtoNet~\cite{snell2017prototypical} & ConvNet64 & $55.50\pm0.70\%$ & $72.00\pm0.60\%$\\
         &MAML~\cite{finn2017model} & ConvNet32 & $58.90\pm1.90\%$ & $71.50\pm1.00\%$\\
         &RENet\cite{DBLP:journals/corr/abs-2108-09666} & ResNet12 & $74.51\pm0.46\%$ & $86.60\pm0.32\%$\\
         &BD-CSPN~\cite{liu2020prototype} & WRN & $72.13\pm1.01\%$ & $82.28\pm0.69\%$\\
         &S2M2\_R~\cite{mangla2020charting} & WRN & $74.81\pm0.19\%$ & $87.47\pm0.13\%$\\
         &PT+NCM~\cite{hu2021leveraging} & WRN & $74.64\pm0.21\%$ & $87.64\pm0.15\%$\\
         &PE$\mathrm{M_b}$E-NCM (ours) & WRN & $\mathbf{74.84\pm0.21}\%$ & $\mathbf{87.73\pm0.15}\%$\\
         \midrule
         \multirow{6}{*}{Transductive}
         &DSN-MR~\cite{simon2020adaptive} & ResNet12 & $78.00\pm0.90\%$ & $87.30\pm0.60\%$\\
         &Transfer+SGC~\cite{hu2021graph} & WRN & $83.90\pm0.22\%$ & $88.76\pm0.15\%$\\
         &MCT~\cite{DBLP:journals/corr/abs-2002-12017} & ResNet12 & $87.28\pm0.70\%$ & $90.50\pm0.43\%$\\
         &PT+MAP~\cite{hu2021leveraging} & WRN & $87.69\pm0.23\%$ & $90.68\pm0.15\%$ \\
         &LST+MAP~\cite{DBLP:journals/corr/abs-2102-05176} & WRN & $87.79\pm0.23\%$ & $90.73\pm0.15\%$ \\
         &PE$\mathrm{M_n}$E-BMS (ours) & WRN & $86.93\pm0.23\%$ & $91.18\pm0.15\%$ \\
         &PE$\mathrm{M_n}$E-BMS$^*$ (ours) & WRN & $\mathbf{87.83\pm0.22\%}$ & $\mathbf{91.20\pm0.15\%}$\\
         \bottomrule
         
    \end{tabular}
    }
    \label{tab:results2}
\end{table}

\begin{table}[h!]
    \caption{1-shot and 5-shot accuracy of state-of-the-art methods when performing cross-domain classification (backbone: WRN).}
    \centering
    \scalebox{0.8}{
    \begin{tabular}{c|l|l|l}
         \toprule
         Setting & Method & 1-shot & 5-shot \\
         \midrule
         \multirow{5}{*}{Inductive}
         &Baseline++~\cite{DBLP:conf/iclr/ChenLKWH19} & $40.44\pm0.75\%$ & $56.64\pm0.72\%$\\
         &Manifold Mixup~\cite{verma2019manifold} & $46.21\pm0.77\%$ & $66.03\pm0.71\%$\\
         &S2M2\_R~\cite{mangla2020charting} & $48.24\pm0.84\%$ & $70.44\pm0.75\%$\\
         &PT+NCM~\cite{hu2021leveraging}  & $48.37\pm0.19\%$ & $70.22\pm0.17\%$\\
         &PE$\mathrm{M_b}$E-NCM (ours) & $\mathbf{50.71\pm0.19}\%$ & $\mathbf{73.15\pm0.16}\%$ \\
         \midrule
         \multirow{5}{*}{Transductive}
         &LaplacianShot~\cite{ziko2020laplacian} & $55.46\%$ & $66.33\%$\\
         &Transfer+SGC~\cite{hu2021graph} & $58.63\pm0.25\%$ & $73.46\pm0.17\%$\\
         &PT+MAP~\cite{hu2021leveraging} & $63.17\pm0.31\%$ & $76.43\pm0.19\%$\\
         &PE$\mathrm{M_n}$E-BMS (ours) & $62.93\pm0.28\%$ & $79.10\pm0.18\%$\\
         &PE$\mathrm{M_n}$E-BMS$^*$ (ours) & $\mathbf{63.90\pm0.31}\%$ & $\mathbf{79.15\pm0.18}\%$\\
         \bottomrule
         
    \end{tabular}
    }
    \label{tab:results_cross}
\end{table}

\subsection{Ablation studies}

\textbf{Generalization to backbone architectures.} To further stress the interest of the ingredients on the proposed method reaching top performance, in Table~\ref{tab:res_backbones} we investigate the impact of our proposed method on different backbone architectures and benchmarks in the transductive setting. For comparison purpose we also replace our proposed BMS algorithm with a standard K-Means algorithm where class prototypes are initialized with the available labelled samples for each class. We can observe that: 1) the proposed method consistently achieves the best results for any fixed backbone architecture, 2) the feature extractor trained on WRN outperforms the others with our proposed method on different benchmarks, 3) there are significant drops in accuracy with K-Means, which stresses the interest of BMS, and 4) the prior on $\mathbf{Q}$ (BMS vs BMS$^*$) has major interest for 1-shot, boosting the performance by an approximation of $1\%$ on all tested feature extractors.

\textbf{Preprocessing impact.} In Table~\ref{tab:res_preprocessing} we compare our proposed PEME with other preprocessing techniques such as Batch Normalization and the ones being used in~\cite{DBLP:journals/corr/abs-1911-04623}. The experiment is conducted on miniImageNet~(backbone: WRN). For all that are put into comparison, we run either a NCM classifier or BMS after preprocessing, depending on the settings. The obtained results clearly show the interest of PEME compared with existing alternatives, we also observe that the power transform helps increase the accuracy on both inductive and transductive settings. We will further study its impact in details.

\begin{table}[h!]
    \caption{1-shot and 5-shot accuracy of proposed method on different backbones and benchmarks. Comparison with k-means algorithm.}
    \centering
    \scalebox{0.57}{
    \begin{tabular}{c|c|c|c|c|c|c|c}
         \toprule
         & & \multicolumn{2}{c|}{\textbf{miniImageNet}} & \multicolumn{2}{c|}{\textbf{CUB}} & \multicolumn{2}{c}{\textbf{CIFAR-FS}}\\
         Method & Backbone & 1-shot & 5-shot & 1-shot & 5-shot & 1-shot & 5-shot \\       
         \midrule
         \multirow{3}{*}{K-MEANS} & ResNet12 & $72.73\pm0.23\%$ & $84.05\pm0.14\%$ & $87.35\pm0.19\%$ & $92.31\pm0.10\%$& $78.39\pm0.24\%$ & $85.73\pm0.16\%$\\ 
         & ResNet18 & $73.08\pm0.22\%$ & $84.67\pm0.14\%$ & $87.16\pm0.19\%$& $91.97\pm0.09\%$& $79.95\pm0.23\%$& $86.74\pm0.16\%$\\
         & WRN & $76.67\pm0.22\%$ & $86.73\pm0.13\%$ & $88.28\pm0.19\%$& $92.37\pm0.10\%$& $83.69\pm0.22\%$& $89.19\pm0.15\%$\\
         \midrule
         \multirow{3}{*}{BMS~(ours)} & ResNet12 & $77.62\pm0.28\%$ & $86.95\pm0.15\%$ & $90.14\pm0.19\%$& $94.30\pm0.10\%$& $81.65\pm0.25\%$ & $88.38\pm0.16\%$\\
         & ResNet18 & $79.30\pm0.27\%$ & $87.94\pm0.14\%$ & $90.50\pm0.19\%$& $94.29\pm0.09\%$& $84.16\pm0.24\%$& $89.39\pm0.15\%$\\
         & WRN & $82.07\pm0.25\%$ & $89.51\pm0.13\%$ & $91.01\pm0.18\%$& $94.60\pm0.09\%$ & $86.93\pm0.23\%$& $91.18\pm0.15\%$\\
         \midrule
         \multirow{3}{*}{BMS$^*$~(ours)} & ResNet12 & $79.03\pm0.28\%$ & $87.01\pm0.15\%$ & $91.34\pm0.19\%$ & $94.32\pm0.09\%$& $82.87\pm0.27\%$ & $88.43\pm0.16\%$\\ 
         & ResNet18 & $80.56\pm0.27\%$ & $87.98\pm0.14\%$ & $91.39\pm0.19\%$ & $94.31\pm0.09\%$ & $85.17\pm0.25\%$ &$89.42\pm0.16\%$\\
         & WRN & $\mathbf{83.35\pm0.25}\%$ & $\mathbf{89.53\pm0.13}\%$ & $\mathbf{91.91\pm0.18}\%$& $\mathbf{94.62\pm0.09}\%$& $\mathbf{87.83\pm0.22}\%$& $\mathbf{91.20\pm0.15}\%$\\
         \bottomrule
         
    \end{tabular}
    }
    \label{tab:res_backbones}
\end{table}

\begin{table}[h!]
    \caption{1-shot and 5-shot accuracy on miniImageNet (backbone: WRN) with different preprocessings on the extracted features.}
    \centering
    \scalebox{0.80}{
    \begin{tabular}{c|c|c|c|c}
         \toprule
         & \multicolumn{2}{c|}{\textbf{Inductive}~(NCM)} & \multicolumn{2}{c}{\textbf{Transductive}~(BMS)} \\
         Preprocessing & 1-shot & 5-shot & 1-shot & 5-shot\\       
         \midrule
         None &   $55.30\pm0.21\%$ & $78.34\pm0.15\%$ & $77.62\pm0.26\%$ & $87.96\pm0.13\%$\\
         Batch Norm~\cite{ioffe2015batch} & $66.81\pm0.20\%$ & $83.57\pm0.13\%$ & $73.74\pm0.21\%$ & $88.07\pm0.13\%$ \\
         L2N~\cite{DBLP:journals/corr/abs-1911-04623} &   $65.37\pm0.20\%$ & $83.46\pm0.13\%$ & $73.84\pm0.21\%$ & $88.15\pm0.13\%$\\
         CL2N~\cite{DBLP:journals/corr/abs-1911-04623} &   $63.88\pm0.20\%$ & $80.85\pm0.14\%$ & $73.12\pm0.28\%$ &$86.47\pm0.15\%$\\
         E$M_b$E&   $68.05\pm0.20\%$ & $83.76\pm0.13\%$ & $80.28\pm0.26\%$ & $88.36\pm0.13\%$\\
         PE$M_b$E&   $\mathbf{68.43\pm0.20}\%$ & $\mathbf{84.67\pm0.13}\%$ & $82.01\pm0.26\%$ &$89.50\pm0.13\%$\\ 
         E$M_n$E& \textbackslash & \textbackslash & $80.14\pm0.27\%$ & $88.39\pm0.13\%$\\
         PE$M_n$E& \textbackslash & \textbackslash & $\mathbf{82.07\pm0.25}\%$ & $\mathbf{89.51\pm0.13}\%$\\
         
         \bottomrule
         
    \end{tabular}
    }
    \label{tab:res_preprocessing}
\end{table}

\textbf{Effect of power transform.} We firstly
conduct a Gaussian hypothesis test on each of the $640$ coordinates of raw extracted features (backbone: WRN) for each of the $20$ novel classes (dataset: miniImageNet). Following D’Agostino and Pearson’s methodology~\cite{diagostino1971omnibus, d1973tests} and $p = 1e-3$,   only one of the $640\times 20 = 12800$ tests return positive, suggesting a very low pass rate for raw features. However, after applying the power transform we record a pass rate that surpasses $50\%$, suggesting a considerably increased number of positive results for Gaussian tests. This experiment shows the effect of power transform being able to adjust feature distributions into more gaussian-like ones.

\begin{figure}[h!]
\begin{center}
\scalebox{0.6}{
    \begin{tikzpicture}
    \begin{axis}[
        xshift=-50mm,
        axis lines = left,
    ]
    \addplot [color=red, ultra thick] coordinates
        {(0.0, 9.0)
(0.01, 22.0)
(0.02, 19.0)
(0.03, 22.0)
(0.04, 26.0)
(0.05, 24.0)
(0.06, 28.0)
(0.07, 23.0)
(0.08, 33.0)
(0.09, 19.0)
(0.1, 28.0)
(0.11, 35.0)
(0.12, 24.0)
(0.13, 12.0)
(0.14, 35.0)
(0.15, 25.0)
(0.16, 12.0)
(0.17, 15.0)
(0.18, 19.0)
(0.19, 15.0)
(0.2, 12.0)
(0.21, 17.0)
(0.22, 8.0)
(0.23, 11.0)
(0.24, 13.0)
(0.25, 18.0)
(0.26, 13.0)
(0.27, 8.0)
(0.28, 7.0)
(0.29, 1.0)
(0.3, 5.0)
(0.31, 5.0)
(0.32, 5.0)
(0.33, 6.0)
(0.34, 5.0)
(0.35, 4.0)
(0.36, 2.0)
(0.37, 2.0)
(0.38, 2.0)
(0.39, 1.0)
(0.4, 2.0)
(0.41, 0.0)
(0.42, 0.0)
(0.43, 1.0)
(0.44, 2.0)
(0.45, 1.0)
(0.46, 0.0)
(0.47, 1.0)
(0.48, 0.0)
(0.49, 0.0)
(0.5, 0.0)};
    \addplot [color=cyan, ultra thick] coordinates
        {(0.0, 60.0)
(0.01, 53.0)
(0.02, 39.0)
(0.03, 41.0)
(0.04, 42.0)
(0.05, 43.0)
(0.06, 27.0)
(0.07, 31.0)
(0.08, 33.0)
(0.09, 26.0)
(0.1, 25.0)
(0.11, 20.0)
(0.12, 15.0)
(0.13, 22.0)
(0.14, 10.0)
(0.15, 6.0)
(0.16, 3.0)
(0.17, 11.0)
(0.18, 10.0)
(0.19, 9.0)
(0.2, 6.0)
(0.21, 5.0)
(0.22, 9.0)
(0.23, 5.0)
(0.24, 7.0)
(0.25, 4.0)
(0.26, 3.0)
(0.27, 5.0)
(0.28, 1.0)
(0.29, 3.0)
(0.3, 3.0)
(0.31, 2.0)
(0.32, 1.0)
(0.33, 2.0)
(0.34, 0.0)
(0.35, 2.0)
(0.36, 1.0)
(0.37, 2.0)
(0.38, 1.0)
(0.39, 1.0)
(0.4, 0.0)
(0.41, 0.0)
(0.42, 0.0)
(0.43, 0.0)
(0.44, 0.0)
(0.45, 0.0)
(0.46, 1.0)
(0.47, 0.0)
(0.48, 2.0)
(0.49, 3.0)
(0.5, 1.0)};
    
    \addplot [color=gray, ultra thick] coordinates
        {(0.0, 79.0)
(0.01, 56.0)
(0.02, 47.0)
(0.03, 43.0)
(0.04, 48.0)
(0.05, 35.0)
(0.06, 39.0)
(0.07, 34.0)
(0.08, 32.0)
(0.09, 31.0)
(0.1, 24.0)
(0.11, 17.0)
(0.12, 14.0)
(0.13, 12.0)
(0.14, 11.0)
(0.15, 9.0)
(0.16, 8.0)
(0.17, 11.0)
(0.18, 7.0)
(0.19, 7.0)
(0.2, 4.0)
(0.21, 3.0)
(0.22, 4.0)
(0.23, 7.0)
(0.24, 2.0)
(0.25, 1.0)
(0.26, 2.0)
(0.27, 5.0)
(0.28, 3.0)
(0.29, 0.0)
(0.3, 0.0)
(0.31, 1.0)
(0.32, 1.0)
(0.33, 1.0)
(0.34, 1.0)
(0.35, 0.0)
(0.36, 0.0)
(0.37, 0.0)
(0.38, 1.0)
(0.39, 0.0)
(0.4, 0.0)
(0.41, 0.0)
(0.42, 0.0)
(0.43, 0.0)
(0.44, 0.0)
(0.45, 0.0)
(0.46, 0.0)
(0.47, 0.0)
(0.48, 0.0)
(0.49, 0.0)
(0.5, 0.0)};
    
    \end{axis}
    \begin{axis}[
        xshift=50mm,
        axis lines = left,
    ]
    \addplot [color=red, ultra thick] coordinates
        {(0.0, 3.0)
(0.01, 5.0)
(0.02, 21.0)
(0.03, 35.0)
(0.04, 45.0)
(0.05, 63.0)
(0.06, 75.0)
(0.07, 77.0)
(0.08, 84.0)
(0.09, 65.0)
(0.1, 61.0)
(0.11, 27.0)
(0.12, 24.0)
(0.13, 7.0)
(0.14, 5.0)
(0.15, 0.0)
(0.16, 2.0)
(0.17, 1.0)
(0.18, 0.0)
(0.19, 0.0)
(0.2, 0.0)
(0.21, 0.0)
(0.22, 0.0)
(0.23, 0.0)
(0.24, 0.0)
(0.25, 0.0)
(0.26, 0.0)
(0.27, 0.0)
(0.28, 0.0)
(0.29, 0.0)
(0.3, 0.0)
(0.31, 0.0)
(0.32, 0.0)
(0.33, 0.0)
(0.34, 0.0)
(0.35, 0.0)
(0.36, 0.0)
(0.37, 0.0)
(0.38, 0.0)
(0.39, 0.0)
(0.4, 0.0)
(0.41, 0.0)
(0.42, 0.0)
(0.43, 0.0)
(0.44, 0.0)
(0.45, 0.0)
(0.46, 0.0)
(0.47, 0.0)
(0.48, 0.0)
(0.49, 0.0)
(0.5, 0.0)};
    \addlegendentry{Class 1}
    \addplot [color=cyan, ultra thick] coordinates
        {(0.0, 16.0)
(0.01, 19.0)
(0.02, 43.0)
(0.03, 50.0)
(0.04, 50.0)
(0.05, 71.0)
(0.06, 60.0)
(0.07, 70.0)
(0.08, 58.0)
(0.09, 45.0)
(0.1, 17.0)
(0.11, 33.0)
(0.12, 25.0)
(0.13, 14.0)
(0.14, 9.0)
(0.15, 5.0)
(0.16, 4.0)
(0.17, 1.0)
(0.18, 7.0)
(0.19, 1.0)
(0.2, 0.0)
(0.21, 0.0)
(0.22, 0.0)
(0.23, 0.0)
(0.24, 1.0)
(0.25, 0.0)
(0.26, 1.0)
(0.27, 0.0)
(0.28, 0.0)
(0.29, 0.0)
(0.3, 0.0)
(0.31, 0.0)
(0.32, 0.0)
(0.33, 0.0)
(0.34, 0.0)
(0.35, 0.0)
(0.36, 0.0)
(0.37, 0.0)
(0.38, 0.0)
(0.39, 0.0)
(0.4, 0.0)
(0.41, 0.0)
(0.42, 0.0)
(0.43, 0.0)
(0.44, 0.0)
(0.45, 0.0)
(0.46, 0.0)
(0.47, 0.0)
(0.48, 0.0)
(0.49, 0.0)
(0.5, 0.0)};
    \addlegendentry{Class 2}
    
    \addplot [color=gray, ultra thick] coordinates
        {(0.0, 16.0)
(0.01, 24.0)
(0.02, 39.0)
(0.03, 43.0)
(0.04, 46.0)
(0.05, 57.0)
(0.06, 66.0)
(0.07, 57.0)
(0.08, 66.0)
(0.09, 59.0)
(0.1, 33.0)
(0.11, 25.0)
(0.12, 25.0)
(0.13, 15.0)
(0.14, 13.0)
(0.15, 9.0)
(0.16, 3.0)
(0.17, 3.0)
(0.18, 1.0)
(0.19, 0.0)
(0.2, 0.0)
(0.21, 0.0)
(0.22, 0.0)
(0.23, 0.0)
(0.24, 0.0)
(0.25, 0.0)
(0.26, 0.0)
(0.27, 0.0)
(0.28, 0.0)
(0.29, 0.0)
(0.3, 0.0)
(0.31, 0.0)
(0.32, 0.0)
(0.33, 0.0)
(0.34, 0.0)
(0.35, 0.0)
(0.36, 0.0)
(0.37, 0.0)
(0.38, 0.0)
(0.39, 0.0)
(0.4, 0.0)
(0.41, 0.0)
(0.42, 0.0)
(0.43, 0.0)
(0.44, 0.0)
(0.45, 0.0)
(0.46, 0.0)
(0.47, 0.0)
(0.48, 0.0)
(0.49, 0.0)
(0.5, 0.0)};
    \addlegendentry{Class 3}
    
    \end{axis}
    \end{tikzpicture}
}
\end{center}
\caption{Distributions of an arbitrarily chosen feature for 3 novel classes before (left) and after (right) power transform.}
\label{fig:analysept}
\end{figure}
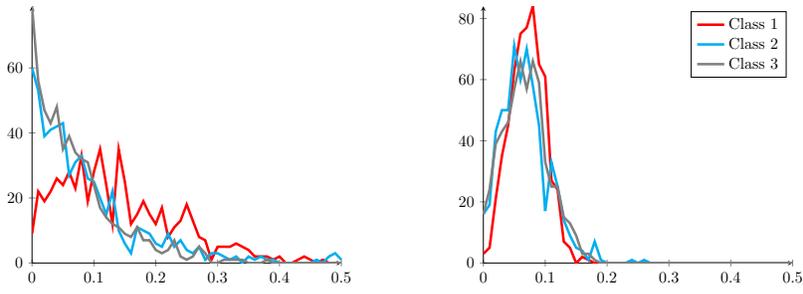

\begin{figure}[h]
\centering
\subfloat{\includegraphics[width=0.42\linewidth]{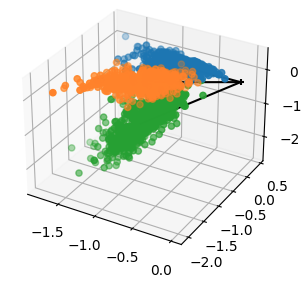}}\qquad
\subfloat{\includegraphics[width=0.45\linewidth]{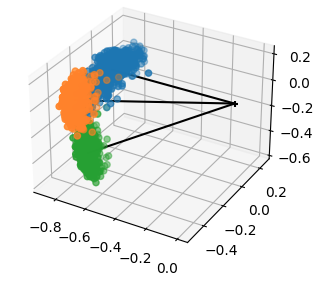}}\\
\caption{Plot of feature vectors (extracted from WRN) from randomly selected 3 classes. (left) Naive features. (right)  Preprocessed features using power transform.}
\label{fig:analysept2}
\end{figure}

To better show the effect of this proposed technique on feature distributions, we depict in Figure~\ref{fig:analysept} the distributions of an arbitrarily selected feature for 3 randomly selected novel classes of miniImageNet when using WRN, before and after applying power transform. We observe quite clearly that 1) raw features exhibit a positive distribution mostly concentrated around 0, and 2) power transform is able to reshape the feature distributions to close-to-gaussian distributions. We observe similar behaviors with other datasets as well. Moreover, in order to visualize the impact of this technique with respect to the position of feature points, in Figure~\ref{fig:analysept2} we plot the feature vectors of randomly selected 3 classes from $\mathbf{D}_{novel}$. Note that all feature vectors in this experiment are reduced to a 3-dimensional ones corresponding to their largest eigenvalues. From Figure~\ref{fig:analysept2} we can observe that power transform, often followed by a L2-normalization, can help shape the class distributions to become more gathered and Gaussian-like~\cite{DBLP:journals/corr/abs-2102-05176}.

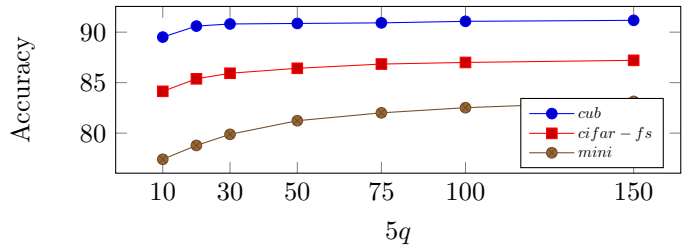
\begin{figure}[h]
  \begin{center}
    \begin{tikzpicture}
       \begin{scope}[]
        \begin{axis}[
            xlabel= $5q$,
            ylabel=Accuracy,
            height=3.8cm,
            width=.75\textwidth,
            xtick = {10, 30, 50, 75, 100, 150},
            legend style={nodes={scale=0.6, transform shape}},
            legend pos={south east,legend cell align=left},
            ]
          
          
          \addlegendentry{$cub$}
          \addplot coordinates
          {(10,89.51) (20,90.60) (30,90.81) (50,90.86) (75,90.92) (100,91.07) (150,91.17)};
          
          \addlegendentry{$cifar-fs$}
          \addplot coordinates
          {(10,84.14) (20,85.38) (30,85.93) (50,86.42) (75,86.84) (100,87.00) (150,87.21)};
          
          \addlegendentry{$mini$}
          \addplot coordinates
          {(10,77.41) (20,78.78) (30,79.88) (50,81.23) (75,82.01) (100,82.52) (150,83.14)};
          
          
        \end{axis}
      \end{scope}
    \end{tikzpicture}
  \end{center}
  \vspace{-.5cm}
  \caption{Accuracy of 5-way, 1-shot classification setting on miniImageNet, CUB and CIFAR-FS as a function of $q$.}
  \label{fig:functionofq}
\end{figure}

\textbf{Influence of the number of unlabelled samples.} In order to better understand the gain in accuracy due to having access to more unlabelled samples, we depict in Figure~\ref{fig:functionofq} the evolution of accuracy as a function of $q$, when the number of classes $n=5$ is fixed. Interestingly, the accuracy quickly reaches a close-to-asymptotical plateau, emphasizing the ability of the method to quickly exploit available information in the task.

\begin{figure}[h]
  \begin{center}
    \begin{tikzpicture}
       \begin{scope}[]
        \begin{axis}[
            xlabel= $e$,
            ylabel=Accuracy,
            height=3.8cm,
            width=.75\textwidth,
            xtick = {0, 10, 20, 30, 40, 50, 60, 70, 80},
            legend style={nodes={scale=0.7, transform shape}, at={(0.65,0.5)}, anchor=west},
            ]
          
          \addlegendentry{BMS$^*$, 1-shot}
          \addplot coordinates
          {(0,82.91) (10,83.24) (20,83.29) (30,82.99) (40,82.43) (50,81.63) (60,80.78) (70,80.03) (80,79.30) };
          
          \addlegendentry{BMS, 5-shot}
          \addplot coordinates
          {(0,88.59) (10,88.99) (20,89.31) (30,89.47) (40,89.51) (50,89.45) (60,89.31) (70,89.19) (80,89.06) };
          
          
        \end{axis}
      \end{scope}
    \end{tikzpicture}
    \begin{tikzpicture}
       \begin{scope}[]
        \begin{axis}[
            xlabel= $\lambda$,
            ylabel=Accuracy,
            height=3.8cm,
            width=.75\textwidth,
            xtick = {3, 4, 5, 6, 7, 8, 9, 10, 12, 15, 20},
            legend style={nodes={scale=0.7, transform shape}, at={(0.65,0.5)}, anchor=west},
            ]
          
          \addlegendentry{BMS$^*$, 1-shot}
          \addplot coordinates
          {(4,82.90) (5,83.21) (6,83.31) (7,83.37) (8,83.36) (9,83.32) (10,83.29) (12,83.09) (15,82.75) (20,82.17)};
          
          \addlegendentry{BMS, 5-shot}
          \addplot coordinates
          {(4,85.56) (5,88.73) (6,89.34) (7,89.50) (8,89.52) (9,89.49) (10,89.41) (12,89.20) (15,88.88) (20,88.53)};
          
          
        \end{axis}
      \end{scope}
    \end{tikzpicture}
  \end{center}
  \vspace{-.5cm}
  \caption{Accuracy of proposed method on miniImageNet (backbone: WRN) as a function of training epoch $e$ (top) and regulation parameter $\lambda$ (bottom).}
  \label{fig:functionofe}
\end{figure}
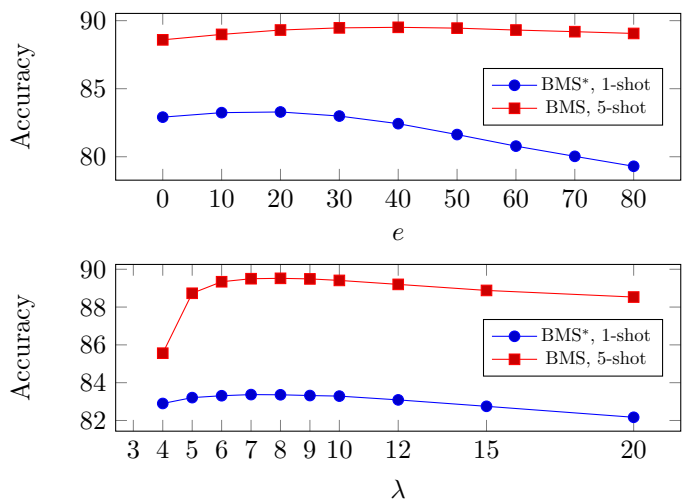

\textbf{Influence of hyperparameters.} In order to test how much impact the hyperparameters could have on our proposed method in terms of prediction accuracy, here we select two important hyperparameters that are used in BMS and observe their impact. Namely the number of training epochs $e$ in logistic regression and the regulation parameter $\lambda$ used for computing the prediction matrix $\mathbf{P}$. In Figure~\ref{fig:functionofe} we show the accuracy of our proposed method as a function of $e$~(top) and $\lambda$~(bottom). Results are reported for BMS$^*$ in 1-shot settings, and for BMS in 5-shot settings. From the figure we can see a slight uptick of accuracy as $e$ or $\lambda$ increases, followed by a downhill when they become larger, implying an overfitting of the classifier. 

\textbf{Proposed method on backbones pre-trained with external data.} In this experiment, we compare our proposed method BMS$^*$ with the work in~\cite{DBLP:journals/corr/abs-2102-03539} that pre-trains the backbone with the help of external illumination data for augmentation, followed by PT+MAP in~\cite{hu2021leveraging} for class center estimation. Here we use the same backbones as~\cite{DBLP:journals/corr/abs-2102-03539}, and replace PT+MAP with our proposed BMS$^*$ at the same conditions. Results are presented in Table~\ref{tab:results_sillnet}. Note that we also show the re-implemented results of~\cite{DBLP:journals/corr/abs-2102-03539}, and our method reaches superior performance on all tested benchmarks using external data in~\cite{DBLP:journals/corr/abs-2102-03539}.

\begin{table}[h]
    \caption{Proposed method on backbones pre-trained with external data. Note that -$re$ denotes the re-implementation of an existing method.}
    \centering
    \scalebox{0.8}{
    \begin{tabular}{l|l|l|l}
         \toprule
         Benchmark & Method & 1-shot & 5-shot  \\       
         \midrule
         \multirow{3}{*}{\textbf{miniImageNet}}
         & Illu-Aug~\cite{DBLP:journals/corr/abs-2102-03539} & $82.99\pm0.23\%$ & $89.14\pm0.12\%$  \\
         & Illu-Aug-$re$ & $83.53\pm0.25\%$ & $89.38\pm0.12\%$ \\
         & PE$M_n$E-BMS$^*$~(ours) & $\mathbf{83.85\pm0.25}\%$ & $\mathbf{90.07\pm0.12}\%$ \\
         \midrule
         \multirow{3}{*}{\textbf{CUB}}
         & Illu-Aug~\cite{DBLP:journals/corr/abs-2102-03539} & $94.73\pm0.14\%$ & $96.28\pm0.08\%$  \\
         & Illu-Aug-$re$ & $94.63\pm0.15\%$ & $96.06\pm0.08\%$ \\
         & PE$M_n$E-BMS$^*$~(ours) & $\mathbf{94.78\pm0.15}\%$ & $\mathbf{96.43\pm0.07}\%$ \\
         \midrule
         \multirow{3}{*}{\textbf{CIFAR-FS}}
         & Illu-Aug~\cite{DBLP:journals/corr/abs-2102-03539} & $87.73\pm0.22\%$ & $91.09\pm0.15\%$  \\
         & Illu-Aug-$re$ & $87.76\pm0.23\%$ & $91.04\pm0.15\%$ \\
         & PE$M_n$E-BMS$^*$~(ours) & $\mathbf{87.83\pm0.23}\%$ & $\mathbf{91.49\pm0.15}\%$ \\
         \bottomrule
         
    \end{tabular}
    }
    \label{tab:results_sillnet}
\end{table}

\textbf{Proposed method on Few-Shot Open-Set Recognition.} Few-Shot Open-Set Recognition~(FSOR) as a new trending topic deals with the fact that there are open data mixed in query set $\mathbf{Q}$ that do not belong to any of the supposed classes used for label predictions. Therefore this often requires a robust classifier that is able to classify correctly the non-open data as well as rejecting the open ones. In Table~\ref{tab:results_sup1} we apply our proposed PEME for feature preprocessing, followed by an NCM classifier, and compare the results with other state-of-the-art alternatives. We observe that our proposed method is able to surpass the others in terms of Accuracy and AUROC.

\begin{table}[h]
    \caption{Accuracy and AUROC of Proposed method for Few-Shot Open-Set Recognition.}
    \centering
    \scalebox{0.7}{
    \begin{tabular}{c|c|c|c|c|c|c|c|c}
         \toprule
         & \multicolumn{4}{c|}{\textbf{miniImageNet}} & \multicolumn{4}{c}{\textbf{tieredImageNet}} \\
         & \multicolumn{2}{c|}{1-shot} & \multicolumn{2}{c|}{5-shot} & \multicolumn{2}{c|}{1-shot} & \multicolumn{2}{c}{5-shot} \\
         Method & Acc & AUROC & Acc & AUROC & Acc & AUROC & Acc & AUROC \\ 
         \midrule
         ProtoNet~\cite{snell2017prototypical} & $64.01\%$ & $51.81\%$ & $80.09\%$ & $60.39\%$ & $68.26\%$ & $60.73\%$ & $83.40\%$ & $64.96\%$ \\
         FEAT~\cite{ye2020few} & $67.02\%$ & $57.01\%$ &$82.02\%$ & $63.18\%$ & $70.52\%$ & $63.54\%$ & $84.74\%$ & $70.74\%$ \\
         NN~\cite{junior2017nearest} & $63.82\%$ & $56.96\%$ &$80.12\%$ & $63.43\%$ & $67.73\%$ & $62.70\%$ & $83.43\%$ & $69.77\%$\\
         OpenMax~\cite{bendale2016towards} & $63.69\%$ & $62.64\%$ &$80.56\%$ & $62.27\%$ & $68.28\%$ & $60.13\%$ & $83.48\%$ & $65.51\%$\\
         PEELER~\cite{liu2020few} & $65.86\%$ & $60.57\%$ &$80.61\%$ & $67.35\%$ & $69.51\%$ & $65.20\%$ & $84.10\%$ & $73.27\%$\\
         SnaTCHer~\cite{jeong2021few} & $67.60\%$ & $70.17\%$ &$82.36\%$ & $77.42\%$ & $70.85\%$ & $74.95\%$ & $85.23\%$ & $82.03\%$\\
         PE$\mathrm{M_b}$E-NCM (ours) & $\mathbf{68.43\%}$ & $\mathbf{72.10}\%$ & $\mathbf{84.67\%}$ & $\mathbf{80.04}\%$ & $\mathbf{71.87\%}$ & $\mathbf{75.44}\%$ & $\mathbf{87.09\%}$ & $\mathbf{83.85}\%$\\
         \bottomrule
    \end{tabular}
    }
    \label{tab:results_sup1}
\end{table}

\subsection{Proposed method on merged features}
In this section we investigate the effect of our proposed method on merged features. Namely, we perform a direct concatenation of raw feature vectors extracted from multiple backbones at the beginning, followed by BMS. In Table~\ref{tab:results_sup2} we chose the feature vectors from three backbones~(WRN, ResNet18 and ResNet12) and evaluated the performance with different combinations. We observe that 1) a direct concatenation, depending on the backbones, can bring about $1\%$ gain in both 1-shot and 5-shot settings compared with the results in Table~\ref{tab:res_backbones} with feature vectors extracted from one single feature extractor. 2) BMS$^*$ reached new state-of-the-art results on few-shot learning benchmarks with feature vectors concatenated from WRN, ResNet18 and ResNet12, given that no external data is used. 

\begin{figure}[h]
  \begin{center}
    \begin{tikzpicture}
       \begin{scope}[]
        \begin{axis}[
            xlabel= $\#$ of backbones,
            ylabel=Accuracy,
            height=3.8cm,
            width=.75\textwidth,
            xtick = {1, 2, 3, 4, 5, 6},
            legend style={nodes={scale=0.6, transform shape}, at={(0.65,0.64)}, anchor=west},
            ]
          
          \addlegendentry{$BMS$, 1-shot}
          \addplot coordinates
          {(1,82.07) (2,83.48) (3,83.80) (4,83.97) (5,84.16) (6,84.31)};
          
          \addlegendentry{$BMS^*$, 1-shot}
          \addplot coordinates
          {(1,83.35) (2,84.65) (3,85.01) (4,85.25) (5,85.43) (6,85.54)};
          
          \addlegendentry{$BMS^*$, 5-shot}
          \addplot coordinates
          {(1,89.53) (2,90.92) (3,91.15) (4,91.34) (5,91.54) (6,91.59)};
          
        \end{axis}
      \end{scope}
    \end{tikzpicture}
  \end{center}
  \vspace{-.5cm}
  \caption{Accuracy of proposed method in different settings as a function of the number of backbones (dataset: miniImageNet).}
  \label{fig:functionofb}
\end{figure}
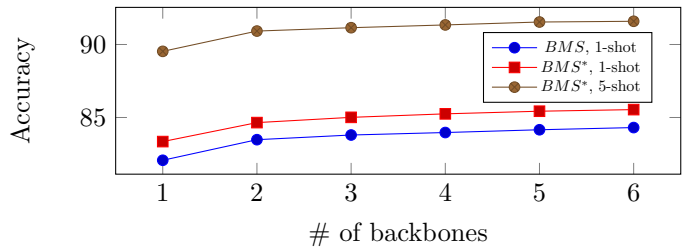

To further study the impact of the number of backbones on prediction accuracy, in Figure~\ref{fig:functionofb} we depict the performance of our proposed method as a function of the number of backbones. Note that here we operate on feature vectors of 6 WRN backbones (dataset: miniImageNet) concatenated one after another, which makes a total of 6 slots corresponding to a $640\times6=3840$ feature size. Each of them is trained the same way as in~\cite{mangla2020charting}, and we randomly select the multiples of $640$ coordinates within the slots to denote the number of concatenated backbones used. The performance result is the average of $100$ random selections and we test with both BMS and BMS$^*$ for 1-shot, and BMS$^*$ for 5-shot. From Figure~\ref{fig:functionofb} we observe that, as the number of backbones increases, there is a relatively steady growth in terms of accuracy in multiple settings of our proposed method, indicating the interest of BMS in merged features.  

\begin{table}[h]
    \caption{1-shot and 5-shot accuracy on miniImageNet, CUB and CIFAR-FS on our proposed PE$M_n$E-BMS with multi-backbones (backbone training procedure follows~\cite{mangla2020charting}, '+' denotes a concatenation of backbone features).}
    \centering
    \scalebox{0.8}{
    \begin{tabular}{l|c|c|c|c|c|c}
         \toprule
         & \multicolumn{2}{c|}{\textbf{miniImageNet}} & \multicolumn{2}{c|}{\textbf{CUB}} & \multicolumn{2}{c}{\textbf{CIFAR-FS}}\\
         Backbone & 1-shot & 5-shot & 1-shot & 5-shot & 1-shot & 5-shot \\       
         \midrule
         RN18+RN12 & $80.32\%$ & $89.07\%$ & $92.31\%$ & $95.62\%$ & $85.44\%$ & $90.58\%$\\
         WRN+RN12 & $82.63\%$ & $90.43\%$ & $92.69\%$ & $95.96\%$ & $87.11\%$ & $91.50\%$\\
         WRN+RN18 & $83.05\%$ & $90.57\%$ & $92.66\%$ & $95.79\%$ & $87.53\%$ & $91.70\%$\\
         WRN+RN18+RN12 & $82.90\%$ & $90.64\%$ & $93.32\%$ & $96.31\%$ & $87.62\%$ & $91.84\%$\\
         WRN+RN18+RN12$^*$ & $84.37\%$ & $90.69\%$ & $\mathbf{94.26}\%$ & $\mathbf{96.32}\%$ & $\mathbf{88.44}\%$ & $\mathbf{91.86}\%$\\
         6$\times$WRN$^*$ & $\mathbf{85.54}\%$ & $\mathbf{91.53}\%$ & \textbackslash & \textbackslash & \textbackslash & \textbackslash\\
         \bottomrule
         \multicolumn{7}{l}{%
            \begin{minipage}{4.5cm}%
            \small $^*$: BMS$^*$. %
            \end{minipage}%
        }
    \end{tabular}
    }
    \label{tab:results_sup2}
\end{table}

\section{Conclusion}
\label{conclusion}

In this paper we introduced a new pipeline to solve the few-shot classification problem. Namely, we proposed to firstly preprocess the raw feature vectors to better align to a Gaussian distribution and then we designed an optimal-transport inspired iterative algorithm to estimate the class prototypes for the transductive setting. Our experimental results on standard vision benchmarks reach state-of-the-art accuracy, with important gains in both 1-shot and 5-shot classification settings. Moreover, the proposed method can bring gains with a variety of feature extractors, with few extra hyperparameters. Thus we believe that the proposed method is applicable to many practical problems. We also provide two versions of our proposed method, one being prior-dependent and one that does not require any knowledge on unlabelled data, and they both are able to bring important gains in accuracy.

\bibliography{mybibfile}

\end{document}